%% file: main.tex
\documentclass[10pt,twocolumn,letterpaper]{article}

\pdfoutput=1

\usepackage{iccv}
\usepackage{times}
\usepackage{epsfig}
\usepackage{graphicx}
\usepackage{amsmath}
\usepackage{amssymb}

\usepackage{url}
\usepackage{mathrsfs}
\usepackage{dsfont}
\usepackage{xspace}
\usepackage{listings}
\usepackage{numprint}
\usepackage{multirow}
\usepackage{setspace}
\usepackage{enumitem}
\usepackage{lipsum}
\usepackage{caption}
\usepackage{booktabs}
\usepackage{algorithm}
\usepackage{algorithmic}
\usepackage{microtype}
\usepackage{sidecap}
\usepackage{subfigure}
\usepackage{balance}
\usepackage{pifont}
\usepackage{caption}
\usepackage{comment}
\usepackage{soul}
\usepackage[accsupp]{axessibility}

\usepackage[dvipsnames]{xcolor}

\usepackage[pagebackref=true,breaklinks=true,letterpaper=true,colorlinks,bookmarks=false]{hyperref}

\iccvfinalcopy 

\def\iccvPaperID{2529} 

\ificcvfinal\fi

\newcommand{\myparagraph}[1]{\vspace{2pt}\noindent{\bf #1}}

\newcommand\blfootnote[1]{%
  \begingroup
  \renewcommand\thefootnote{}\footnote{#1}%
  \addtocounter{footnote}{-1}%
  \endgroup
}

\begin{document}

\title{
Generalized and Incremental Few-Shot Learning by\\
Explicit Learning and Calibration without Forgetting }

\author{Anna Kukleva\textsuperscript{1} \ \ \ \ \ \
Hilde Kuehne\textsuperscript{2,3} \ \ \ \ \ \
Bernt Schiele\textsuperscript{1} \\

\small \textsuperscript{1}MPI for Informatics, Saarland Informatics Campus \
\small \textsuperscript{2}CVAILab, Goethe University Frankfurt \
\small \textsuperscript{3} MIT-IBM Watson AI Lab, Cambridge
}

\maketitle
\ificcvfinal\thispagestyle{empty}\fi

\begin{abstract}
Both generalized and incremental few-shot learning have to deal with three major challenges:
learning novel classes from only few samples per class,
preventing catastrophic forgetting of base classes, and
classifier calibration across novel and base classes.
In this work we propose a three-stage framework that allows to explicitly and effectively address these challenges.
While the first phase learns base classes with many samples, the second phase learns a calibrated
classifier for novel classes from few samples while also preventing catastrophic forgetting.
In the final phase, calibration is achieved across all classes.
We evaluate the proposed framework on four challenging benchmark datasets for image and video few-shot classification and obtain state-of-the-art results for both generalized and incremental few shot learning.

\blfootnote{https://github.com/annusha/LCwoF}
\end{abstract}

\vspace{-5mm}
\input{1_intro}
\input{2_rel_work}
\input{3_method}
\input{4_results}
\input{5_conclusion}

{\small
\bibliographystyle{ieee_fullname}
\bibliography{main}
}

\newpage
{\normalsize
\appendix
\input{6_supmat}

}

\end{document}

%% file: 1_intro.tex
\section{Introduction}
\label{sec:intro}
In this paper we are interested in two practically important learning scenarios, namely generalized few-shot learning (GFSL) ~\cite{gidaris2018dynamic, ren2019incremental, qi2018imprinted, shi2019relational} and incremental few-shot learning (IFSL) \cite{tao2020few, chen2021incremental}. In both scenarios it is possible to learn a performant classifier for a set of base classes for which many training samples exist. However, for the novel classes, only few training samples are available such that a \textit{novel class learning} is challenging. Additionally, in generalized few-shot learning and in  incremental learning it is important to prevent \textit{catastrophic forgetting} of the base classes during novel class learning. Last, but not least, \textit{classifier calibration} across classes has to be addressed, due to the imbalance in the amount of training samples.
While previous work focuses on addressing a subset of these challenges~\cite{gidaris2018dynamic, ren2019incremental, kumar2019protogan, tao2020few, chen2021incremental}, in this paper we aim to address all  three.

\begin{figure}
\centering
\includegraphics[width=0.95\linewidth]{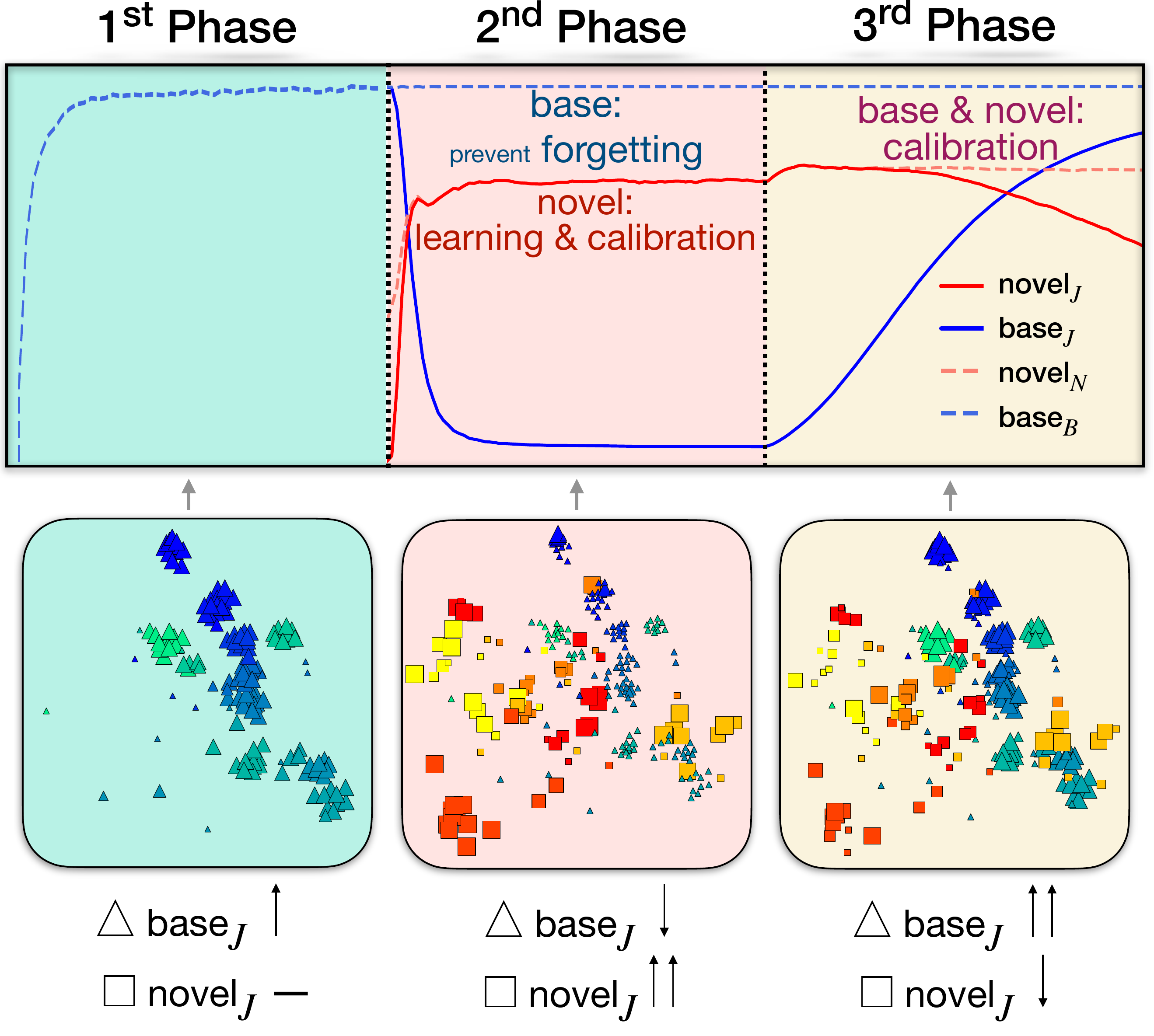}
\caption{Overview of the performance of our framework during different phases.
\textit{J} indicates performance in the joint space, \textit{B} and \textit{N} denote performance in the base and novel spaces respectively.
During the $1^{st}$ phase we train the model on the base classes.
During the $2^{nd}$ phase we try to achieve a high performance on novel classes in the joint space and prevent forgetting of base classes.
In the $3^{rd}$ phase we calibrate the two classifiers and achieve balanced performance between base and novel classes. T-SNE plots show the performance of the test samples at each phase. The symbol size shows the confidence of the model for the sample.}
\label{fig:teaser}
\vspace{-5.5mm}
\end{figure}

To this end, we propose a three phase framework to explicitly address these challenges.
The first phase is devoted to general representation learning as in previous work~\cite{gidaris2018dynamic, ren2019incremental, wang2018low}.
Here, we utilize a large base dataset for pretraining and obtain high performance 
for base classification.
In the second phase we concentrate on learning novel classes. In contrast to the prior work, we pay special attention
to training a calibrated classifier for the novel classes while simultaneously preventing catastrophic forgetting for the base classes.
More specifically we propose base-normalized cross entropy that amplifies the softmax output of novel classes to overcome the bias towards the base classes, and simultaneously enforce the model to preserve previous knowledge via explicit weight constraints.
In the third phase we address the problem of calibrating the overall model across base and novel classes.
In Fig.~\ref{fig:teaser} we show how the model develops during all three phases by plotting the test accuracy of base and novel classes in the separate and joint spaces.
%
%
The contributions of this work are as follows:
\begin{enumerate}[wide, label={\textbf{(\arabic*)}}, itemsep=-.3em,
topsep=0em, labelwidth=0em, labelindent=0pt]
    \item A framework to explicitly address the problems of generalized few-shot-learning by balancing between learning novel classes, forgetting base classes and calibration across them in three phases;
    \item Base-normalized cross-entropy 
    to overcome the bias learned by the model on the base classes in combination with weight constraints to mitigate the forgetting problem in the second phase; 
    \item An extensive study to evaluate the proposed framework on images and videos showing state-of-the-art results for generalized and incremental few shot learning.
\end{enumerate}



%% file: 2_rel_work.tex
\section{Related Work}
\label{sec:rel_work}

\myparagraph{Generalized Few-Shot Learning (GFSL)}
Many approaches for few-shot learning (FSL) rely on a meta-learning paradigm to quickly adapt a method to new underrepresented samples~\cite{snell2017prototypical, vinyals2016matching, finn2017maml, sung2018learning, mishra2018snail, sung2018learning}. Such models can be hard to extend to a generalized setup since they do not explicitly learn classification of base classes and do not consider extreme imbalance. Some recent work on few-shot learning additionally examine a generalized setup~\cite{ye2020set2set, luo2019global} showing a significant drop in performance in the joint space.

First work~\cite{hariharan2017low, wang2018low} on GFSL propose to hallucinate extra samples based on intra-class variations of base classes. Later, Gidaris~\etal~\cite{gidaris2018dynamic} propose an attention based weights generator for few-shot classes and promote cosine normalization
to unify recognition of base and novel classes. Concurrently, Qi~\etal~\cite{qi2018imprinted} propose weight imprinting that is also based on the idea of cosine normalization.  The technique is widely used to avoid the explicit calibration of magnitudes~\cite{hou2019learning, ren2019incremental, shi2019relational, mittal2021essentials, gidaris2019generating, rebuffi2017icarl}.
In our work we exploit the bias in base classification weights to give the model impetus to learn novel classes in the joint space.
Ren~\etal~\cite{ren2019incremental} propose to use an attention attractor network~\cite{zemel2001localist} to regularize the learning of novel classes with respect to accumulated base attractor vectors. The above works are based on meta-learning frameworks, consequently they can be dependent on the number of novel classes. In contrast, Shi~\etal~\cite{shi2019relational} propose a graph-based framework to model the relationship between all  classes that can be trained end-to-end.
GFSL receives attention in the video domain as well. Previous work propose to enlarge the training data for few novel classes by means of a generative adversarial network~\cite{kumar2019protogan} or to retrieve similar data from a large-scale external dataset~\cite{xian2020generalized}.

\begin{figure*}[ht!]
\centering
\includegraphics[width=0.95\linewidth]{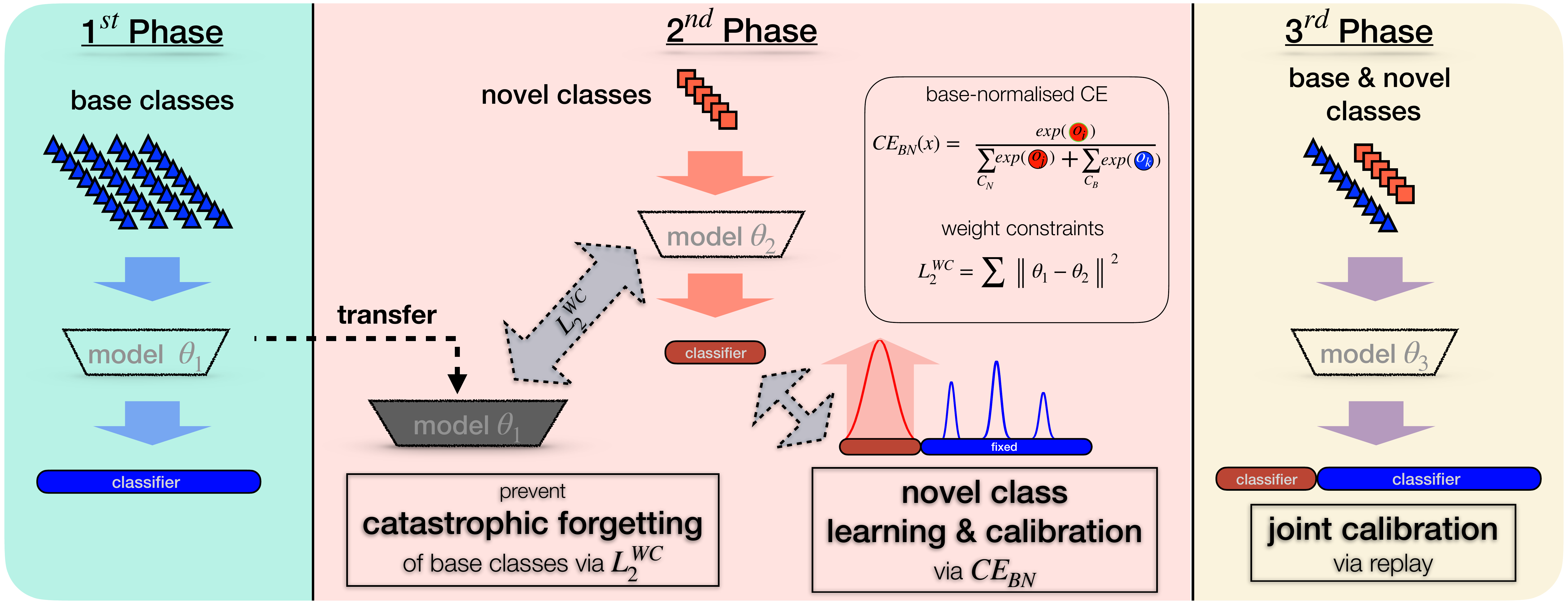}
\caption{Overview of our framework. To achieve balanced performance on base and novel classes we deal with three problems \textit{learning novel classes}, \textit{catastrophic forgetting}, and \textit{calibration} that we address in different phases of our framework. In the $1^{st}$ phase we pretrain model on base classes with abundant data. During the $2^{nd}$ phase we employ $L_2^{WC}$ weight constraints to preserve knowledge and base-normalized cross entropy ($CE_{BN}$) to calibrate learning of novel classes in the joint space with base classes. In the $3^{rd}$ phase we calibrate the performance with the balanced replay of novel and base samples.
}
\label{fig:fig2}
\vspace{-3mm}
\end{figure*}

\myparagraph{Incremental Few-Shot Learning (IFSL)}
Class incremental learning (CIL)~\cite{li2017learningwf, liu2020mnemonics, castro2018end, hou2019learning, rebuffi2017icarl} addresses the problem of a continuously growing classification space, where each set of novel classes extends previously observed classes. 
The major problem of incremental learning, catastrophic forgetting~\cite{mccloskey1989catastrophic}, is caused by limiting the access to already seen data while each novel class is provided with a large train set.
Due to the ample number of training samples, on the one hand, authors~\cite{mittal2021essentials, ahn2020ssil, zhao2020continual} propose to separate softmax classification into several subspaces to balance learning. On the other hand, some work addresses it with bias removal techniques~\cite{wu2019large, belouadah2019il2m, Belouadah_2020_scail} by training batch of additional hyperparameters~\cite{wu2019large}, using dual memory~\cite{belouadah2019il2m} or with post-processing~\cite{Belouadah_2020_scail}. On the contrary, we benefit from the joint space to overcome the deficiency of data and to learn a stronger classifier for novel classes without introducing any additional parameters.

A new task that combines FSL and CIL is incremental few-shot learning (IFSL)~\cite{chen2021incremental, tao2020few, ayub2020cognitively}. GFSL can be seen as a subproblem of IFSL with only one incremental learning set of novel classes. Tao~\etal~\cite{tao2020few} propose to preserve the topology of the feature manifold via a neural gas network.
Chen~\etal~\cite{chen2021incremental} use a non-parametric method based on learning vector quantization in deep embedding space to avoid imbalanced weights in a parametric classifier. In this work we approach the problem from the perspective of classic parametric classification models~\cite{he2016deep, krizhevsky2012imagenet} that recently are shown to be effective in FSL~\cite{tian2020rethink} and CIL~\cite{prabhu2020gdumb}.


%% file: 3_method.tex

\vspace{-1mm}
\section{LCwoF-Framework}
\label{sec:method}
In the following we introduce the general setting and the motivation of our method based on four separate performance measures introduced below. We then discuss the second phase training as the most crucial part to achieve strong performance for both novel and base classes, followed by our third phase training. Finally, we discuss how to generalize our method to incremental few shot learning.

\myparagraph{Setting:}
In both generalized and incremental few-shot learning we have a set of base classes $C_B$ with many training samples. Additionally, we have one or several sets of novel classes $C_N$ with only few training samples.
In generalized few-shot learning we have just one set of novel classes, while in incremental few-shot learning we have a sequence of such sets. In the following, to keep the notation simple, we discuss our approach based on a single set of novel classes, whereas in incremental learning the approach is applied to the sequence of such sets of novel classes.
Note that incremental few-shot learning in our work is the same as the few-shot class incremental learning in~\cite{chen2021incremental, tao2020few}.

\begin{figure}
\centering
\vspace{-6mm}
\includegraphics[width=0.7\linewidth]{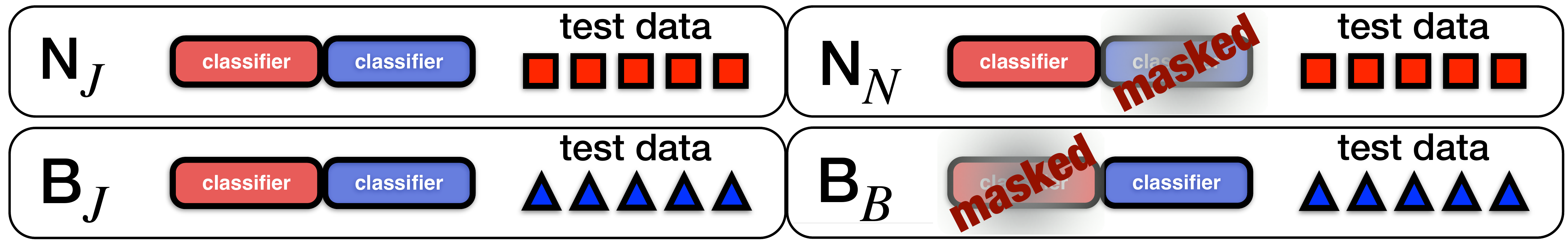}
\vspace{-3mm}
\caption{Classification layers for different evaluation protocols.}
\label{fig:classifiers_explained}
\vspace{-5mm}
\end{figure}

\myparagraph{Performance measures and approach:}
In few-shot learning we are interested to achieve best performance for both base and novel classes simultaneously.
Therefore, in order to monitor performance for both sets of classes, we are considering four different measures (see Fig.~\ref{fig:classifiers_explained})
First, we denote $B_{/B}$ the classification performance of \textit{base} samples in the space of only \textit{base} classes $C_B$,
and $N_{/N}$ the classification performance of \textit{novel} samples in the space of only \textit{novel} classes $C_N$.
More importantly, we are interested in the performance in the \textit{joint} (J) space where both base and novel classes are accounted for simultaneously $C_B \cup C_N$.
For this we consider the performance of base and novel samples separately in the \textit{joint} space, that is $N_{/J}$ and  $B_{/J}$.
We prefer these two measures rather than using only the joint performance in joint space due to the imbalance of the number of
classes~\cite{hou2019learning, belouadah2019il2m, wu2019large} with $|C_B| \gg |C_N|$ (e.g. 64 base vs. 5 novel).

These measures are directly related to the three challenges mentioned above:
novel class learning is measured by $N_{/N}$ and $N_{/J}$,
catastrophic forgetting by $B_{/B}$ and $B_{/J}$,
while calibration is related to $B_{/J}$ and $N_{/J}$.
While ideally we would like to address all the measures simultaneously, we found this to be difficult in practice.
Instead, during the first phase of our framework, we optimize for $B_{/B}$.
In the second phase, during novel class learning, we are aiming for a calibrated classifier for the novel classes and thus optimize for both $N_{/N}$ and $N_{/J}$,
instead of only $N_{/N}$ as in standard few-shot learning.
Simultaneously, we aim to prevent catastrophic forgetting by an additional weight regularization that keeps $B_{/B}$ high (see Fig.~\ref{fig:teaser}).
In the third and last phase we aim to calibrate across novel and base classes and thus optimize for both $B_{/J}$ and $N_{/J}$.

\vspace{-1mm}
\myparagraph{Model parameters:}
We denote the backbone parameters as $\theta_1$, $\theta_2$, and $\theta_3$ for the first, the second, and the third phase respectively. As classifier we use a linear classification layer without bias that we train on the top of the backbone. Practically, during the second phase we introduce a classification layer for novel classes. To evaluate performance in the joint space we concatenate the output of the two classifiers before the normalization. $o_i$ denotes the output logit of the model for the classification into class $i$. We train $\theta_1$ on a large dataset of base classes to obtain a good representation. For the second and the third phase we initialize $\theta_{phase}$ with the parameters $\theta_{phase-1}$ and fine-tune on the corresponding to phase set.

\subsection{Second Phase - Novel Class Training}

\myparagraph{Base-Normalized Cross Entropy ($\pmb{CE_{BN}}$)}
Recently, Tian~\etal~\cite{tian2020rethink} showed that in few-shot learning competitive classification on novel classes can be achieved given good representations using standard cross entropy without meta-learning and prototypes. We follow this idea and train $\theta_2$ using a pretrained model $\theta_1$ from the first phase and fine-tune it with a new classification layer for novel classes.
The standard way to fine-tune the model on the training set that includes $C_N$ classes is
\begin{equation}
\label{eq:CE}
CE(x) = \sum_{C_N}y_i ln\Big(\frac{exp(o_i)}{\sum\limits_{C_N}exp(o_j)}\Big),
\end{equation}
where $o_i$ is the logit of the corresponding class $i \in C_N$, and $y_i$ equals to one if $x$ belongs to class $i$, otherwise to zero.

One problem here is that even if the model is capable of learning information about the new classes $C_N$ well, there is no guarantee that this performance is replicated in the joint space $N_{/J}$. By training two disjoint classifiers we learn classification weights that satisfy the classification problem either on novel classes $N_{/N}$ or on base classes $B_{/B}$, but so far the model does not learn any correlation between base classification weights and novel classification weights.

To this end, we propose to provide the model with information about the base class distribution in the joint space using readily available information. Specifically, for each novel training sample we compute logits not only for the novel $C_N$ classes, but also for the base $C_B$ classes (note, in the second phase the base classes classifier is kept fixed).
We use these logits to compute classification scores with the softmax function, thus the normalization of each score includes the base class logits that initially prevail in the sum, as follows:
\begin{equation}
\label{eq:CE_BN}
 CE_{BN}(x) = \sum_{C_N}y_i ln\Big(\frac{exp(o_i)}{\sum\limits_{C_N}exp(o_j) + \sum\limits_{C_B}exp(o_k)}\Big).
\end{equation}
%
With this normalization, the novel model learns output probabilities for the novel classes directly in the joint space, and specifically increasing magnitude of novel class logits with respect to base class logits. This allows to have a good classification accuracy for novel classes $N_{/N}$ in the second phase and at the same time helps to match this accuracy in the joint space $N_{/J}$.
Note that we do not use any base class training samples 
during this learning phase and we keep the weights of the base classifier fix.

\myparagraph{Knowledge Preservation}
After the first phase the model performs well for the base classes and we aim to keep this capability.
In FSL~\cite{Dhillon2020Abaselinefew, raghu2019rapid} and  IL~\cite{li2017learningwf, rebuffi2017icarl} multiple works show that adaptive representations can be beneficial for learning novel classes, specifically to fine-tune the parameters of the representation.
In IL, the typical way to preserve knowledge from base classes~\cite{castro2018end, lee2019overcoming, wu2019large, douillard2020podnet, li2017learningwf, rebuffi2017icarl} is knowledge distillation (KD)~\cite{hinton2015distilling} that is applied in the form of KL-divergence between logits of base classes from adapted and old models.

As an alternative to keep the network to remember about the previous knowledge, we propose to utilise explicit weight constraints (WC) of the model with respect to the old model from the first phase. We formulate it in form of a $L_2$ penalization over adaptive parameters of the representation~\cite{li2017learningwf, kirkpatrick2017overcoming, evgeniou2004regularized}:
\begin{equation}
\label{eq:WC_2}
L_2^{WC} = \sum \left\lVert \theta_{1} - \theta_{2} \right\lVert ^2,
\end{equation}
where $\theta_2$ denotes adaptive parameters of the backbone excluding classification parameters during the second phase and $\theta_1$ are the parameters of the model after base pretraining.
The above constraint forces the model to keep the representation learned on base samples, but still allows the model to adjust the weights of the representation to better fit novel classes while not diverging a lot from the old model.
The overall loss for the second phase is thus:
\begin{equation}
\label{eq:2phase}
L = CE_{BN} + \lambda L_2^{WC},
\end{equation}
where $\lambda$ controls the strength of knowledge preservation.

\subsection{Third Phase - Joint Calibration}
\myparagraph{Balanced Replay}
The first and the second phase account for the performance on the base classes ($B_{/B}$) and  for the novel class learning in both spaces ($N_{/N}$ and $N_{/J}$). For the third phase, due to the difference of number in training samples for base and novel classes and preservation of $B_{/B}$ during the second phase, the model is able to obtain good performance in the joint space as well. Empirically we found that during the second phase $B_{/J}$ performance can drop drastically, but due to keeping the base class performance $B_{/B}$ we can achieve good $B_{/J}$ performance in the third phase.

To achieve a balanced performance in the joint space of base and novel classes we apply the replay technique that is common in incremental learning~\cite{rebuffi2017icarl, liu2020mnemonics, rajasegaran2020itaml, lee2019overcoming}. Specifically, we randomly draw only once base training samples, one per class, and join these samples with the novel training data. Moreover, in our case we require the least possible memory~\cite{rebuffi2017icarl} to store exemplars of base classes, an essential component for incremental learning.

We continue training the model on the balanced dataset in the joint space. Due to the initial strong bias towards base classes from the first phase ($B_{/B}$), the model can improve its performance for base classes in the joint space ($B_{/J}$) quickly while at the same time overwriting the novel class performance at least partially ($N_{/J}$).

\subsection{From Generalized to Incremental Learning}

The main difference between GFSL~\cite{ren2019incremental, shi2019relational} and IFSL~\cite{tao2020few, chen2021incremental} is the number of few-shot tasks.
So far we considered the case of GFSL and it can be regarded as the first two tasks in terms of incremental setting: the training of the base classes refers to task one, the training of novel classes to task two in the incremental setup.
As the current framework addresses the joint generalized problem in three phases, base classes, novel classes, and joint classes, we can easily extend the architecture to more tasks by repeating the novel class training.
Specifically, for each new few-shot task we apply the second phase to learn a good joint classification for the current classes.

To evaluate the performance of the current joint space we finalize the training with the last phase of recuperation.
To this end, we keep exemplars from base and novel classes from different tasks and perform training with the base-normalized cross-entropy loss and $L_2^{WC}$ weights constraints as before.
So, each time when we need to evaluate the joint performance on all classes, we apply the third phase.

For example, to report accuracy after five tasks, we learn representation parameters from base classes in the first task, as next stage we then apply the second phase sequentially for the second, third, fourth, and fifth tasks respectively, each time enlarging the classifier by the number of new classes in the task.
After the fifth task we apply the third phase, balanced replay, where for each few-shot task we use all available data, and one sample per class for the data from the first task.
During test the performance of the model is evaluated on a set that contains all previously seen classes.

%% file: 4_results.tex
\input{sota_mini}

\input{sota_tiered}

\section{Experimental Results}
\label{sec:results}
This section validates our proposed LCwoF-framework. First, we compare our method to the previous state-of-the-art work on both GFSL and IFSL in Section~\ref{sec:sota}. Then we analyze each phase and the components separately to show the importance and connections of each to the improved performance in Section~\ref{sec:ablations}.

\myparagraph{Datasets:}
\textit{mini-ImageNet}~\cite{vinyals2016matching} is a 100-class subset of  ImageNet~\cite{deng2009imagenet}. For FSL we follow~\cite{ren2019incremental} and use a subsets 64-12-24 classes that corresponds to base-val-novel classes, and for IFSL we follow~\cite{chen2021incremental, tao2020few} with a subset of 60 and 40 classes for base training and incremental few-shot testing with 5 classes per each novel set.  
\textit{tiered-ImageNet}~\cite{ren2018meta} is a larger subset of Imagenet~\cite{deng2009imagenet} with categorical splits for for base, validation, and novel classes. Here, each high-level category (e.g. dog that includes different breeds) belongs only to one of the splits. 
\textit{mini-Kinetics}~\cite{xian2020generalized} is a 100-class subset of the Kinetics video classification dataset~\cite{Kay2017Kinetics}. We use the splits from~\cite{xian2020generalized}. 
\textit{UCF101}~\cite{Soomro2012UCF101} is a video dataset with 101 classes in total. We follow~\cite{kumar2019protogan} with a splitting of 50-51 for base and novel classes for FSL on videos. We additionally introduce and evaluate more challenging division of the dataset.

\vspace{1mm}
\myparagraph{Implementation details}
For the FSL experiments on mini-ImageNet and tiered-ImageNet we employ the same ResNet12 with DropBlock~\cite{ghiasi2018dropblock}  as backbone and pretrain it on base classes for 500 epochs with SGD optimizer with momentum with the learning rate (lr) of 1e-3 that is decayed by 0.1 at 75, 150, and 300 epochs. For the second and the third phase we use lr of 1e-2 and 1e-3 respectively for the classification layers and decayed by 0.1 lr for the backbone parameters. For IFSL experiments we use ResNet18 and follow the same pretraining steps as above.  We use different architecture choices for GFSL and IFSL to remain comparable to previous works after the base pretraining. For the second phase we always train the model for 150 epochs, while for the third phase we use validation set to choose the number of epochs for each dataset.
For videos we preextract features with C3D model~\cite{tran2015learning} pretrained on large-scale Sports-1M~\cite{karpathy2014large} dataset. We apply average pooling over temporal domain to obtain one feature vector per video. As a backbone we use 2-layer MLP. We also clip gradients at value 100 for the experiments. More details can be found in the supplement.
\myparagraph{Evaluation}
We evaluate the proposed framework primarily with respect to the harmonic mean ($hm_{/J}$)~\cite{schonfeld2019generalized, shi2019relational, kumar2019protogan} that is computed between base and novel performance in the joint space. Additionally, we report performance of base and novel classes in their respective subspaces ($B_{/B}$ and $N_{/N}$), in the joint space ($B_{/J}$ and $N_{/J}$), and the arithmetic mean over the joint space ($am_{/J}$) as in~\cite{ren2019incremental}. Extended tables for all datasets are in the supplement. 5w1s and 5w5s denote 5 novel classes with 1 and 5 training samples per classes respectively. For the state-of-the-art comparison, we average over 600 episodes~\cite{shi2019relational, luo2019global, Dhillon2020Abaselinefew}, for all other experiments over 100 episodes. \textit{unlim} denotes access to the entire base training set, whereas for \textit{lim} setup we use small subset. All specifications are in supplement.

\subsection{Comparison to state-of-the-art}
\label{sec:sota}
\vspace{-1mm}
\myparagraph{Generalized Few-Shot Learning}
We compare our performance on image and video benchmarks: mini-ImageNet, tiered-ImageNet, Kinetics and UCF101 in Tables~\ref{table:mini_sota_5w1s},~\ref{table:tiered_sota_5w1s},~\ref{table:kinetics_sota_5w1s}, and~\ref{table:ucf_sota_50w1s} respectively.
For mini-ImageNet and tiered-ImageNet, we train the respective backbones from scratch on the base classes. For Kinetics and UCF101, we preextract video features as in~\cite{kumar2019protogan, xian2020generalized} and then train a shallow MLP model on the base classes.

On mini-ImageNet in Table~\ref{table:mini_sota_5w1s} we provide an evaluation in separate and joint space on 5w1s and 5w5s setups. For both backbones, conv4~\cite{shi2019relational} and ResNet, we achieve significant improvements over state-of-the-art results in terms of $hm_{/J}$. Here, we can observe that previous methods drop in performance on both novel ($N_{/N}$) and base ($B_{/B}$) classes, whereas we address the problem by explicitly balancing  between forgetting, learning, and calibration and achieve better performance.

On tiered-ImageNet, Table~\ref{table:tiered_sota_5w1s}, we can observe a similar pattern and achieve strong improvements. Here, even with more base classes, we are able to calibrate the performance between novel and base classes.

We further evaluate the performance of the proposed idea on two video datasets. Our results on Kinetics, shown in Table~\ref{table:kinetics_sota_5w1s}, and UCF101, shown in Table~\ref{table:ucf_sota_50w1s}, show that the proposed framework is able to perform well on the pre-extracted features.
Results on UCF101 we present on two different splits for training and testing. In Table~\ref{table:ucf_sota_50w1s} the first two lines correspond to splits provided by~\cite{kumar2019protogan} thus can be directly compared.
The second part of the table shows the evaluation for the setup with the original UCF train/test split as defined in \cite{Soomro2012UCF101}.

Note that on both image datasets we obtain significant gains in performance while applying \textit{unlim} sampling strategy, while on the Kinetics and UCF101 there is a slight decrease in comparison to \textit{lim}. We speculate that it happens due to the fixed feature preextraction whereas on images we train models on raw images.
Across all the datasets, setting and architectures we consistently achieve significantly better performance than previous work.

\input{sota_kinetics}
\input{sota_ucf}

\input{sota_incremental}
\myparagraph{Incremental Few-Shot Learning}
We compare our method with the current few-shot class incremental methods in  Table~\ref{table:incremental_sota_hm}. As in the previous experiments we use  the hm accuracy that we compute between base (first set) and novel classes.
We provide more detail on the performance of each task in the supplement, specifically showing performance of base and novel classes separately, as well as the standard accuracy of all classes in the joint space. Our method notably outperforms other methods in the field due to the fact that we address directly the balance between the performance on base and novel classes. We show that we obtain higher novel accuracy in the joint space for every task.

\subsection{Ablation Studies}
\label{sec:ablations}
Here we investigate the influence of several components of our framework and  the impact of various hyperparameters.

\begin{figure}
\centering
\includegraphics[width=\linewidth]{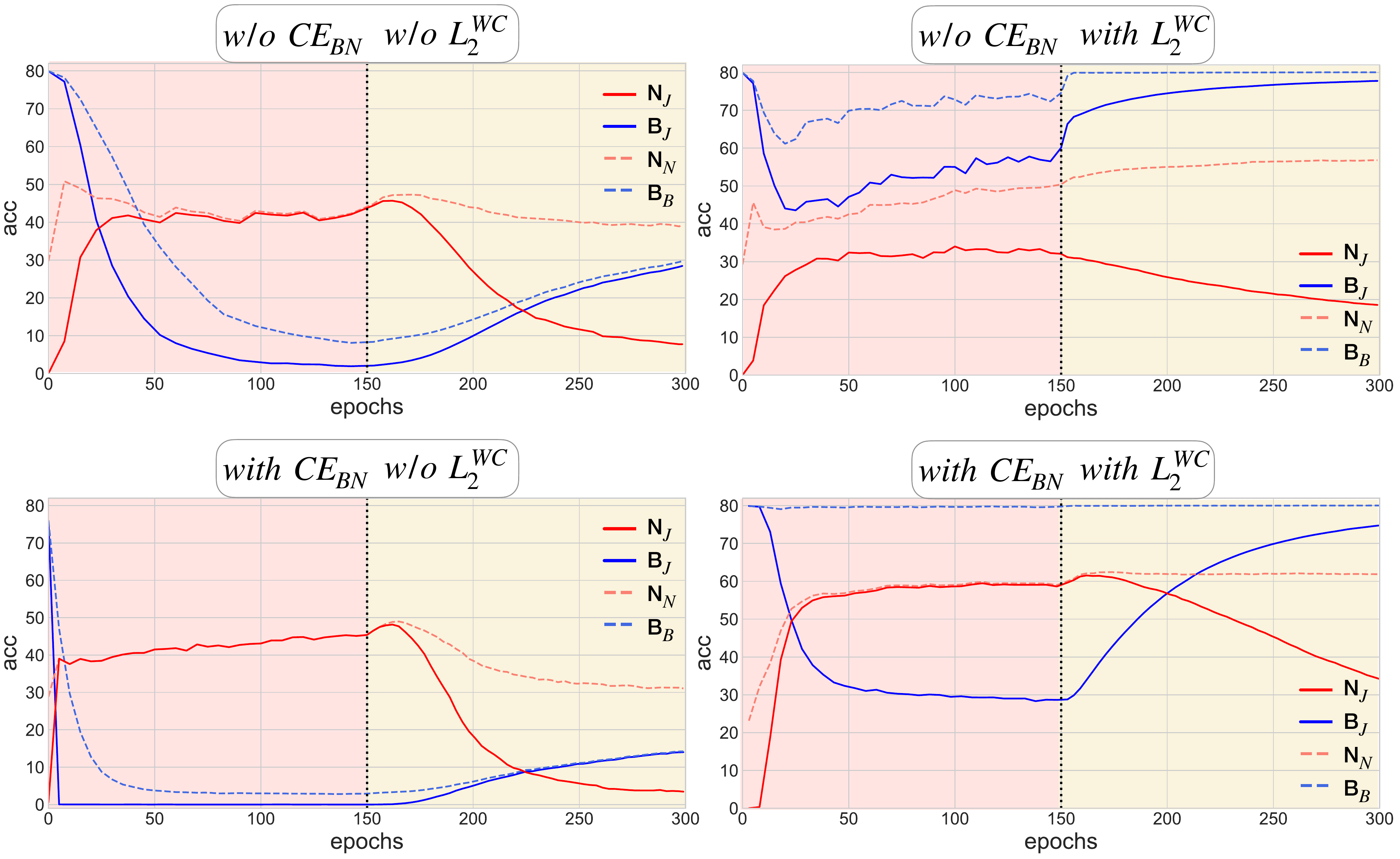}
\caption{Influence of the proposed $CE_{BN}$ and $L_2^{WC}$ on the training during the $2^{nd}$ and the $3^{rd}$ phase. Red lines correspond to novel classes, blue to base classes. Solid lines represent performance in the joint space, dashed lines in separate corresponding subspaces. \textit{Top left}: base performance drops as $B_{/B}$ and $B_{/J}$ and is not able to recover during the $3^{rd}$ phase; \textit{bottom left}: both $N_{/N}$ and $N_{/J}$ achieve high performance, base performance still is not able to recover; \textit{top right}:  $L_2^{WC}$ helps to recuperate base performance in the joint space while novel performance drops; \textit{bottom right}: by addressing both $N_{/J}$ with $CE_{BN}$ and $B_{/B}$ with $L_2^{WC}$ we allow the model recover during the $3^{rd}$ phase on $B_{/J}$ while not losing drastically on novel classes.}
\label{fig:gce_l2}
\vspace{-2mm}
\end{figure}

\myparagraph{Calibration and forgetting during the second phase}
In this subsection we analyse the influence of the base-normalized cross entropy as a technique to address the calibration problem as well as the influence of knowledge preservation to address the forgetting issue. In Fig.~\ref{fig:gce_l2} we show the behaviour of the training process during the second and the third phase with and without base-normalized cross entropy and knowledge preservation as explicit weight constraints $L_2^{WC}$.
Removing both elements, as shown in the top left sub-figure, results in a drastic drop in $B_{/B}$ and $B_{/J}$ that prohibits the model to quickly recover during the replay (third) phase since all previous knowledge is lost.
The left bottom sub-figure shows the impact of $CE_{BN}$. With $CE_{BN}$ the model easily achieves good performance on $N_{/J}$ in the joint space that matches $N_{/N}$, but here both $B_{/B}$ and $B_{/J}$ drop during the second phase training that again prevents  recuperation.
Compared to that, the right top sub-figure includes base knowledge preservation that keeps both $B_{/B}$ and $B_{/J}$ relatively high, and further facilitates complete recovery for the base classes. But without $CE_{BN}$ novel learning in the joint space suffers during both the second and the third phase.
The right bottom sub-figure shows performance if both, $CE_{BN}$ and knowledge preservation, are used. The model is able to keep a certain performance of $B_{/B}$ during the second phase while achieving high accuracy on the novel classes ($N_{/N}$ and $N_{/J}$). During the third phase we calibrate the model by the replay technique and achieve good balance between the two disjoint sets of classes.

\input{distill_loss}
\myparagraph{Knowledge Preservation}
One important factor of the method is that we aim to retain the knowledge that the model obtained in the first phase, specifically the $B_{/B}$ performance, during the second phase. In this section we evaluate two different methods to achieve this objective, comparing the proposed explicit weight constraints $L_2^{WC}$ with knowledge distillation that is formulated via $KL$-divergence.
Knowledge distillation is a common technique to preserve knowledge in incremental learning~\cite{castro2018end, douillard2020podnet, lee2019overcoming, li2017learningwf, rebuffi2017icarl, wu2019large}, where abundant training data is available for new classes.

In Table~\ref{table:distill_loss} we evaluate the performance on 100 episodes on mini-ImageNet for the 5w1s and 5w5s settings. Comparing $L_2^{WC}$ and $KD$ knowledge preservation techniques, we find that the latter marginally outperforms the other if more data is available, as in the 5w5s setting, whereas plain weight constraints are more efficient in the lowest data regime with 1 training sample per class (5w1s). Additionally, we evaluate $KD$ by including additional unlabeled 1000 images from the validation set during the second phase for $KD$ loss computations, denoted as $KD^+$ in Table~\ref{table:distill_loss}. We find that it helps to improve in the 5w1s setting, but still stays lower than $L_2^{WC}$. Applying both techniques at the same time does not give an improvement.

\begin{figure}
\centering
\includegraphics[width=\linewidth]{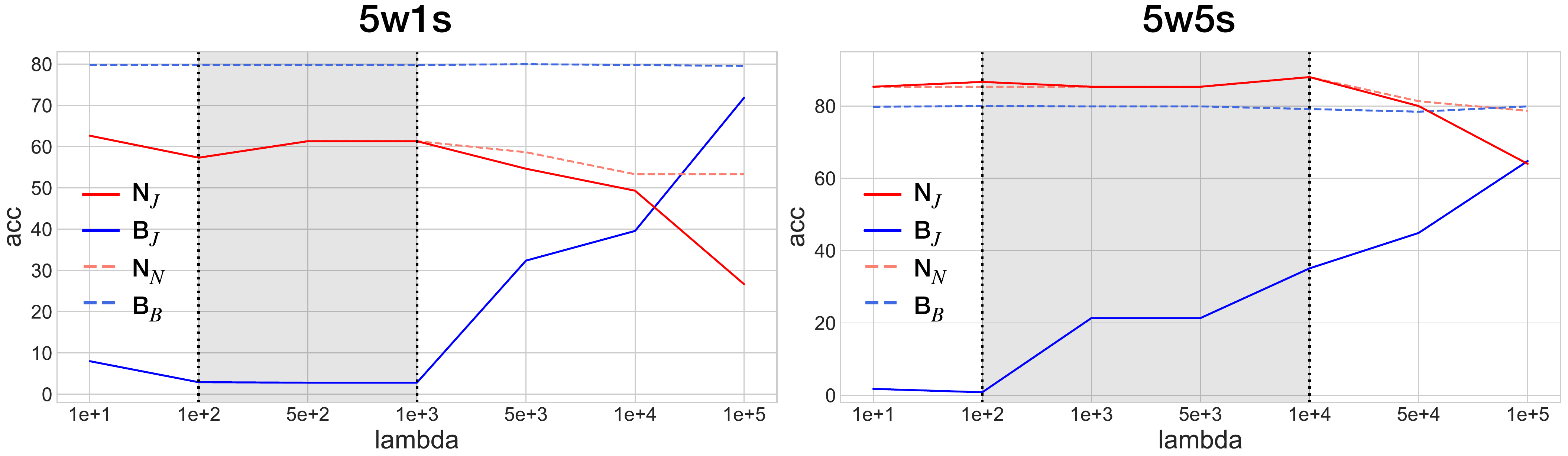}
\caption{Evaluation of the performance after the $2^{nd}$ phase of the framework on the base and novel classes that depends on importance weight $\lambda$ of the $L_2^{WC}$ term. Filled areas correspond appropriate $\lambda$ that allows us to achieve the desirable performance (before $N_{/J}$ drops). Results on mini-ImageNet. }
\label{fig:l2_lambda}
\vspace{-1mm}
\end{figure}

\myparagraph{Impact of $\lambda$}
In this section we study the influence of $\lambda$ for the $L_2^{WC}$ loss and how to preserve more knowledge and drop less on the base classes $B_{/J}$, but also how the model behaviour changes if we enforce it to preserve even more.
In Fig.~\ref{fig:l2_lambda} we plot the accuracy after training in the second phase, before the replay phase for different $\lambda$. In Fig.~\ref{fig:l2_lambda} we fill these areas with gray color, that allow to reach our two objectives for the second phase, i.e., achieving good performance in the joint space on novel classes $N_{/J}$ and  keeping good accuracy in the base space $B_{/B}$. Specifically higher $\lambda$ helps to further keep $B_{/B}$ performance while novel learning $N_{/N}$ and $N_{/J}$ starts decreasing with higher values. It shows the start of the decrease depends on the number of available training samples: the more training data, the more we can keep from base classes by choosing a higher $\lambda$.

\input{early_stop_2nd_phase}
\myparagraph{Early Stopping: Second Phase}
We observe that during the second phase, usually $B_{/J}$ starts dropping even when $N_{/J}$ has already achieved some reasonable accuracy, as can e.g. be seen in the bottom right subplot of Figure~\ref{fig:gce_l2}.
We, therefore, also report the best performance that can be achieved with different $\lambda$ during the second phase in Table~\ref{table:early_stop_2nd_phase}.
Note that $\lambda$ influences the contribution of the knowledge preservation part $L_2^{WC}$. Thus, $B_{/J}$ will drop faster if a lower $\lambda$ is chosen. By adjusting $\lambda$, we find that the proposed technique can also be helpful during the second phase. It shows that in this case, balanced model performance can be reached with higher $\lambda$ and that we can achieve high performance even without the third replay phase.

\input{freeze_batch_norm}
\myparagraph{Impact of batch normalization}
 We use batch norm layers in the model that capture statistic from the base classes during the first phase. At the second phase our data is highly limited thus we fix batch norm during further training. Table~\ref{table:freeze_batch_norm} shows that the performance drops more than 1 point when the model tries to accumulate new statistic from 1 training sample per class and to adapt parameters respectively.

\myparagraph{Impact of normalization} One of the common strategies to unify magnitudes of base and novel classifiers is to use cosine normalization of the embedding and weights~\cite{hou2019learning, ren2019incremental, shi2019relational, mittal2021essentials, gidaris2019generating, rebuffi2017icarl}. We experiment with such a setup (Table~\ref{table:cos_bias}, lines 1 \& 3) for our framework and find a decline in performance for both, $N_{/J}$ and $B_{/J}$.
Note that the performance of the model after the first phase on $B_{/B}$ is the same as without cosine normalization.
But if we attempt to match $N_{/J}$ and $N_{/N}$ during the second phase, we find that constrained magnitudes of logits due to normalization restrict the performance and do not allow to achieve our second phase objectives.

\myparagraph{Analysis of classification layer}
As default, we conduct all our experiments with a  linear classification layer without bias term (Table~\ref{table:cos_bias}, lines 2 \& 3). We therefore assess the performance of the model with and without bias. We find that during the first phase, it is the same, but  that during the second and the third phase it is beneficial to use the latter giving a boost of about 0.7 in the performance $hm_{/J}$.

\input{bias_cos}

%% file: sota_mini.tex

\setlength\belowrulesep{0pt}
\setlength\aboverulesep{0pt}

\begin{table*}[!th]
\centering
\footnotesize
\tabcolsep=0.15cm
\begin{tabular}{c|cc|cc|cc||cc|cc|cc}  
\toprule
& \multicolumn{6}{c}{mini-ImageNet 5w1s} & \multicolumn{6}{c}{mini-ImageNet 5w5s} \\
\multirow{2}{*}{method} &  $N_{/N}$ & $B_{/B}$ & $N_{/J}$  & $B_{/J}$  & \multirow{2}{*}{$hm_{/J}$} & \multirow{2}{*}{$am_{/J}$} &  $N_{/N}$ & $B_{/B}$ & $N_{/J}$  & $B_{/J}$  & \multirow{2}{*}{$hm_{/J}$} & \multirow{2}{*}{$am_{/J}$} \\
 & ($5_{/5}$) & ($64_{/64}$) & ($5_{/69}$) & ($64_{/69}$) &  &  & ($5_{/5}$) & ($64_{/64}$) & ($5_{/69}$) & ($64_{/69}$) &  &  \\
 \midrule
CONV4 & & &&&&&&&&&& \\
 \midrule
 PN~\cite{snell2017prototypical}$^{\circ}$ & 53.88 & 54.02 & 0.02 & 54.02 & 0.04 & 27.02 & 70.84 & 60.42 & 2.99 & 60.41 & 5.70 & 31.70 \\
 DFSL~\cite{gidaris2018dynamic}$^{\circ}$ & 55.80 & 69.93 & 40.30 & 58.54 & 47.74 & 49.42 & 72.24 & 70.24 & 58.26 & 59.89 & 59.06 & 59.07 \\
 RGFSL~\cite{shi2019relational} & 55.08 & 65.14 & 39.86 & 54.65 & 46.10 & 47.25 & 72.32 & 67.79 & 56.32 & 59.30 & 57.71 & 57.81 \\ 
 LCwoF (ours) \textit{lim} & 58.32 & 72.75  & 47.16 & 55.07 & \textbf{50.81} & \textbf{51.12}  & 73.63 & 71.82 & 62.23 & 59.94 & \textbf{61.06} & \textbf{61.09} \\
 \midrule 
 ResNet & & &&&&&&&&&& \\
 \midrule
  PN~\cite{snell2017prototypical}$^{\star}$ & - & - & - & - & - & 42.73& - & - & - & - & - & 57.05  \\
   IW~\cite{qi2018imprinted}(i) & 47.17 & 61.78 & 31.25 & 47.72 & 37.77 & 39.49 & 67.56 & 69.07 & 46.96 & 58.92 & 52.26 & 52.94 \\
 DFSL~\cite{gidaris2018dynamic} (c) & 56.83 & 70.15 & 41.32 & 58.04 & 48.27 & 49.68  &72.82 & 70.03 & 59.27 & 58.68 & 58.97 & 58.98 \\
 AAN~\cite{ren2019incremental}(c) & 56.14 & 77.58 & 45.61 & 63.92 & 53.24 & 54.76 &69.72 & 77.58 & 60.82 & 64.14 & 62.43 & 62.48 \\ 
  AAN~\cite{ren2019incremental}(orig) & - & - & - & - & - & 54.95 & - & - & - & - & - & 63.04 \\
 LCwoF (ours) \textit{lim} & 60.78 & 79.89 & 53.78 & 62.89 & \textbf{57.39} & \textbf{57.84} & 77.65 & 79.96 & 68.58 & 64.53 & \textbf{66.49} & \textbf{66.55} \\
 LCwoF (ours) \textit{unlim} & 61.15 & 80.10 & 53.33 & 62.99 & \textbf{57.75} & \textbf{58.16} & 77.88 & 80.09 & 67.17 & 66.59 & \textbf{66.88} & \textbf{66.88} \\
\bottomrule
\end{tabular}

\caption{Comparison to state-of-the-art on mini-ImageNet 5w1s (left) and 5w5s (right) with backbones CONV4 and ResNet. \textit{lim} denotes limited access to base train samples during the third phase, for \textit{unlim} we do not apply such restrictions.  $^{\circ}$ indicates results copied from RGFSL~\cite{shi2019relational}, $^{\star}$ indicates results from AAN~\cite{ren2019incremental}, (c) denotes that we run available code on the corresponding data, (i) states for our re-implementation of the respective method, (orig) indicates original numbers from the respective paper.}
\label{table:mini_sota_5w1s}
\vspace{-2mm}
\end{table*}

%% file: sota_tiered.tex
\begin{table}[!t]
\vspace{1mm}
\centering
\footnotesize
\tabcolsep=0.15cm
\begin{tabular}{c|cc|c||cc|c}  
\toprule

& \multicolumn{3}{c}{tiered-ImageNet 5w1s} & \multicolumn{3}{c}{tiered-ImageNet 5w5s} \\
method & $N_{/J}$  & $B_{/J}$  & $hm_{/J}$ & $N_{/J}$  & $B_{/J}$  & $hm_{/J}$ \\
 \midrule
 

 
 IW~\cite{qi2018imprinted}(i) & 44.95 & 62.53 & 52.30 & 71.85 & 56.11 & 63.01 \\
 DFSL~\cite{gidaris2018dynamic}(c) & 47.32 & 36.10 & 40.96 & 67.94 & 39.08 & 49.61 \\
 AAN~\cite{ren2019incremental}(c) & 54.39 & 55.85 & 55.11 & 57.76 & 64.13 & 64.93 \\ 
 LCwoF \textit{lim} & 57.13 & 60.39 & \textbf{58.71} & 69.05 & 63.44 & \textbf{66.12} \\
 LCwoF \textit{unlim} & 59.79 & 60.86 & \textbf{58.75} & 70.20 & 63.01 & \textbf{66.41} \\

\bottomrule
\end{tabular}
\caption{Comparison to state-of-the-art on tiered-ImageNet 5w1s (left) and 5w5s (right) with ResNet backbone. $N_{/J}$ equal to $5_{/205}$, $B_{/J}$ to $200_{/205}$. \textit{lim}, \textit{unlim}, (i) and (c) see in Table~\ref{table:mini_sota_5w1s}. \\ }
\label{table:tiered_sota_5w1s}
\vspace{-5mm}
\end{table}


 



%% file: sota_kinetics.tex

\begin{table}[!t]
\centering
\footnotesize
\tabcolsep=0.15cm
\begin{tabular}{c|cc|c||cc|c}  
\toprule
& \multicolumn{3}{c}{mini-Kinetics 5w1s} & \multicolumn{3}{c}{mini-Kinetics 5w5s} \\
method & $N_{/J}$  & $B_{/J}$  & $hm_{/J}$ & $N_{/J}$  & $B_{/J}$  & $hm_{/J}$  \\
 \midrule
 

 
 IW~\cite{qi2018imprinted}(i) & 45.56 & 48.56 & 47.01 & 56.92 & 49.17 & 52.76 \\
 DFSL~\cite{gidaris2018dynamic}(c) & 50.81 & 44.51 & 47.45 & 70.29 & 46.31 & 55.83 \\
 GFSV~\cite{xian2020generalized} & 13.70 & 88.70 & 23.73 & 22.30 & 88.70 & 35.64 \\
 ANN~\cite{ren2019incremental}(c) & 46.13 & 35.96 & 40.41 & 56.99 & 43.21 & 49.15 \\ 
 LCwoF \textit{lim} & 47.51 & 50.84 & \textbf{49.12} & 63.65 & 54.55 & \textbf{58.75} \\
 LCwoF \textit{unlim} & 46.26 & 51.94 & \textbf{48.93} & 65.40 & 52.70 & \textbf{58.37} \\

\bottomrule
\end{tabular}

\caption{Comparison to state-of-the-art on mini-Kinetics 5w1s (left) and 5w5s (right) with MLP backbone on pre-extracted features. $N_{/J}$ equal to  $5_{/69}$, $B_{/J}$ to $64_{/69}$. \textit{lim}, \textit{unlim}, (i) and (c) see in Table~\ref{table:mini_sota_5w1s}.}
\label{table:kinetics_sota_5w1s}
\vspace{-5mm}
\end{table}






%% file: sota_ucf.tex

\begin{table}[!t]
\centering
\footnotesize
\tabcolsep=0.15cm
\begin{tabular}{c|cc|c}  
\toprule
& \multicolumn{3}{c}{UCF101 50w1s} \\
method &  $N_{/J}$  & $B_{/J}$  & $hm_{/J}$ \\
 \midrule
 ProtoG~\cite{kumar2019protogan} & 52.30 & 75.30 & 61.72 \\
  LCwoF (ours) \textit{lim} & 54.41 & 91.41 & \textbf{68.22} \\
 
 \midrule
 
 IW~\cite{qi2018imprinted}(i) & 45.22 & 76.15 & 56.74 \\
 LCwoF (ours) \textit{lim} & 50.78 & 70.72 & \textbf{59.11} \\
 LCwoF (ours) \textit{unlim} & 49.12 & 69.98 & \textbf{57.72} \\

\bottomrule
\end{tabular}

\caption{Comparison to state-of-the-art on UCF101 50w1s with MLP backbone on pre-extracted features. $N_{/J}$ equal to $50_{/101}$, $B_{/J}$ to $51_{/101}$. \textit{lim}, \textit{unlim} and (i) see in Table~\ref{table:mini_sota_5w1s}. }
\label{table:ucf_sota_50w1s}
\vspace{-3mm}
\end{table}

%% file: sota_incremental.tex

\begin{table*}[!ht]
\centering
\begin{tabular}{ll}
\centering
\footnotesize
\tabcolsep=0.15cm
\begin{tabular}{c|cccccccc}  
\toprule
\multicolumn{9}{c}{mini-ImageNet 5w5s} \\
method  & 2 & 3 & 4 & 5 & 6 & 7 & 8 & 9 \\
$hm_{/J}$ & $+5$ & $+10$ & $+15$ & $+20$ & $+25$ & $+30$ & $+35$ & $+40$  \\
\midrule
FT$^\diamond$ & 7.23 & 7.39 & 4.87 & 2.40 & 2.06 & 1.84 & 1.57 & 1.40  \\
Joint$^\diamond$ & 8.92 & 17.02 & 21.86 & 20.54 & 22.92 & 22.85 & 24.41 & 24.95  \\
\midrule
iCaRL~\cite{rebuffi2017icarl}$^\diamond$  & 8.45 & 13.86 & 14.92 & 13.00 & 14.06 & 12.74 & 12.16 & 11.71  \\
UCIR~\cite{hou2019learning}$^\diamond$ & 9.62 & 14.14 & 15.58 & 13.19 & 13.63 & 13.11 & 12.76 & 11.96  \\
PN~\cite{snell2017prototypical}$^\diamond$ & 9.76 & 14.72 & 16.78 & 19.09 & 20.06 & 19.37 & 18.98 & 18.90  \\
ILVQ~\cite{xu2012ilvq}$^\diamond$ & 9.66 & 16.08 & 17.78 & 20.05 & 20.35 & 19.64 & 19.06 & 18.89  \\
SDC~\cite{yu2020sdc}$^\diamond$  & 20.51 & 18.79 & 17.36 & 20.47 & 19.21 & 18.27 & 20.79 & 21.77  \\
IW~\cite{qi2018imprinted}$^\diamond$  & 25.32 & 20.45 & 22.62 & 25.48 & 22.54 & 20.66 & 21.27 & 22.27  \\
IDLVQ~\cite{chen2021incremental} & 21.69 & 20.44 & 21.98 & 25.19 & 22.99 & 20.82 & 21.56 & 22.65  \\
\midrule
LCwoF \textit{lim} & \textbf{41.24} & \textbf{38.96} & \textbf{39.08} & \textbf{38.67} & \textbf{36.75} & \textbf{35.47} & \textbf{34.71} & \textbf{35.02}  \\

\bottomrule
\end{tabular}
&
\centering
\scriptsize
\tabcolsep=0.15cm
\begin{tabular}{c|cccc}  
\toprule
\multicolumn{5}{c}{mini-ImageNet 5w5s}  \\
$B_{/J}$ ($60$) & 2 & 5 & 7 & 9 \\
\midrule
Joint$^\diamond$  & 63.30 &  62.18 & 61.86  & 61.89  \\

SDC~\cite{yu2020sdc}$^\diamond$ & 63.58  & 60.29  & 59.05  & 59.87  \\
IW~\cite{qi2018imprinted}$^\diamond$ & 63.52 & 61.17  & 60.63  & 59.64  \\
IDLVQ~\cite{chen2021incremental} & 63.77  & 61.22  & 60.97 & 60.44  \\
\midrule
LCwoF \textit{lim}  & 57.33 & 51.38  & 47.60 & 47.73  \\

\midrule
\midrule
$N_{/J}$ & 2 & 5 & 7 & 9 \\
$\#cl$& $5$ & $20$ & $30$ & $40$  \\
\midrule
Joint$^\diamond$ & 4.80 & 12.30 & 14.01 & 15.62  \\

SDC~\cite{yu2020sdc}$^\diamond$  & 12.23 &  12.33  & 10.81  & 13.30  \\
IW~\cite{qi2018imprinted}$^\diamond$  & 15.81 &  16.09 & 12.45 & 13.69  \\
IDLVQ~\cite{chen2021incremental}  & 13.07 & 15.86  & 12.55  & 13.94  \\
\midrule
LCwoF \textit{lim}  & \textbf{32.20} & \textbf{31.12}  & \textbf{28.27} & \textbf{27.65}  \\

\bottomrule
\end{tabular}

\end{tabular}

\caption{IFSL. Comparison to state-of-the-art on mini-ImageNet. Number of base classes for the first task is 60. Each next task increases joint space by 5 novel classes with 5 training samples per class.  $^\diamond$ indicates results copied from IDLVQ~\cite{chen2021incremental}. Left: $hm_{/J}$ between base and novel classes in the joint space for each task; right top: base classification accuracy in the joint space; right bottom: novel classification accuracy in the joint space.}
\label{table:incremental_sota_hm}

\vspace{-3mm}
\end{table*}

%% file: distill_loss.tex

\setlength\belowrulesep{1pt}
\setlength\aboverulesep{2pt}

\begin{table}[!t]
\centering
\small
\tabcolsep=0.15cm
\begin{tabular}{c|ccc|ccc}  
\toprule
& \multicolumn{3}{c|}{5w1s} & \multicolumn{3}{c}{5w5s} \\
\multirow{2}{*}{method}  & $N_{/J}$  & $B_{/J}$  & \multirow{2}{*}{$hm_{/J}$} & $N_{/J}$ & $B_{/J}$  & \multirow{2}{*}{$hm_{/J}$} \\
&  ($5/_{69}$) & ($64/_{69}$) &  & ($5/_{69}$) & ($64/_{69}$) &  \\

\midrule
$L_2^{WC}$ & 53.28 & 63.24 & \textbf{57.83} & 68.61 & 64.73 & 66.61 \\
$KD$ & 50.28 & 57.72 & 53.74 & 70.29 & 64.67 & \textbf{67.36} \\
$KD^+$ & 45.87 & 67.20 & 54.52 & 68.01 & 65.88 & 66.93 \\
$L_2^{WC} + KD$ & 53.43 & 60.51 & 56.75 & 68.45 & 63.47 & 65.86 \\
\bottomrule
\end{tabular}

\caption{Comparison of knowledge preservation techniques, such as $L_2^{WC}$ as explicit weights constraints, knowledge distillation ($KD$) for old classes during the $2^{nd}$ phase, $KD^+$  includes $KD$ on the additional 1000 unlabeled images during the second phase, and combination $L_2^{WC} + KD$. Results are computers on mini-ImageNet. 
}

\label{table:distill_loss}
\vspace{-5mm}
\end{table}

%% file: early_stop_2nd_phase.tex

\begin{table}[!t]
\centering
\small
\tabcolsep=0.15cm
\begin{tabular}{c|ccc}  
\toprule
& \multicolumn{3}{c}{5w1s}  \\
$\lambda$  & $N_{/J}$  & $B_{/J}$  & $hm_{/J}$ \\
\midrule
1e+4 & 50.89 & 50.68 & 50.78 \\
5e+4 & 53.06 & 58.56 & 55.67  \\
1e+5 & 52.65 & 58.70 & 55.51  \\
\midrule
LCwoF & 53.28 & 63.24 & \textbf{57.83}  \\

\bottomrule
\end{tabular}

\caption{Best possible performance of the model that can be achieved during the second phase (without the third phase). $\lambda$ stands for the importance weight of the $L_2^{WC}$ term. Higher $\lambda$, higher knowledge preservation, higher accuracy $B_{/J}$. We apply GCE and $L_2^{WC}$ during the second phase with various $\lambda$. Results on mini-ImageNet.
}
\label{table:early_stop_2nd_phase}
\end{table}

%% file: freeze_batch_norm.tex

\begin{table}[!t]
\centering
\small
\tabcolsep=0.15cm
\begin{tabular}{c|ccc}  
\toprule
 & $N_{/J}$  & $B_{/J}$  & $hm_{/J}$ \\
\midrule
no & 54.39 & 59.13 & 56.66  \\
yes & 53.28 & 63.24 & \textbf{57.83} \\

\bottomrule
\end{tabular}

\caption{Influence of freezing batch norm during the $2^{nd}$ and the $3^{rd}$ phases, results on mini-ImageNet 5w1s.}
\label{table:freeze_batch_norm}
\vspace{-3mm}
\end{table}

%% file: bias_cos.tex

\begin{table}[!t]
\centering
\small
\tabcolsep=0.15cm
\begin{tabular}{lc|ccc}  
\toprule
& & $\text{N}_{/J}$  & $\text{B}_{/J}$  & $\text{hm}_{/J}$ \\
\midrule
1 & cos.norm. & 47.91 & 59.69 & 53.15  \\
2 & bias & 52.96 & 62.05 & 57.14 \\
3 & no bias & 53.28 & 63.24 & \textbf{57.83} \\

\bottomrule
\end{tabular}

\caption{Comparison between different linear layers: cos is cosine normalization of the weights and the embedding space, bias stands for linear layer with the bias term, no bias stands for linear layer without the bias term. Results on mini-ImageNet 5w1s.}
\label{table:cos_bias}
\vspace{-5mm}
\end{table}

%% file: 5_conclusion.tex
\section{Conclusion}
\label{sec:conclusion}
This paper addresses major challenges in generalized few-shot and incremental few-shot learning with our three-phase framework. First, we learn a powerful representation by training a model on base classes. In the second phase, concerned with novel class learning, we employ base-normalized cross entropy that calibrates novel class classifiers to overcome the bias towards base classes.
Additionally, during that phase we preserve knowledge about base classes via weight constraints. In the third phase, to achieve calibrated classifiers across both base and novel classes, we employ balanced replay.
We show that each phase of the framework allows to study and address the essential problems of the task explicitly. We evaluate our proposed framework on four benchmark image and video datasets and achieve state-of-the-art performance across all settings.
This work can be seen as a first step towards more explicitly addressing
calibration, learning and knowledge preservation jointly to further improve
deep learning for imbalanced settings beyond the ones addressed in this paper.

%% file: 6_supmat.tex
In the supplement in Section~\ref{sec:abl_sup} we provide additional ablation experiments on the second and the third phase, further in Section~\ref{sec:impl_det_sup} we expand implementation details by specifying the applied data augmentation, generalized and incremental few-shot learning setups, and the splits used for UCF101 dataset.
Finally, Section~\ref{sec:tabels_sup} contains extended tables for all the datasets.

\section{Additional ablation of the phases}
\label{sec:abl_sup}
In this section we discuss possible variations of our proposed framework and their influence on the performance.
Specifically, we discuss the necessity of the second phase and the duration of the second phase.
We also inspect the influence of knowledge preservation on the performance after the third phase and the impact of weight decay regularization.

\subsection{The second phase}
\begin{figure}[ht]
\centering
\includegraphics[width=\linewidth]{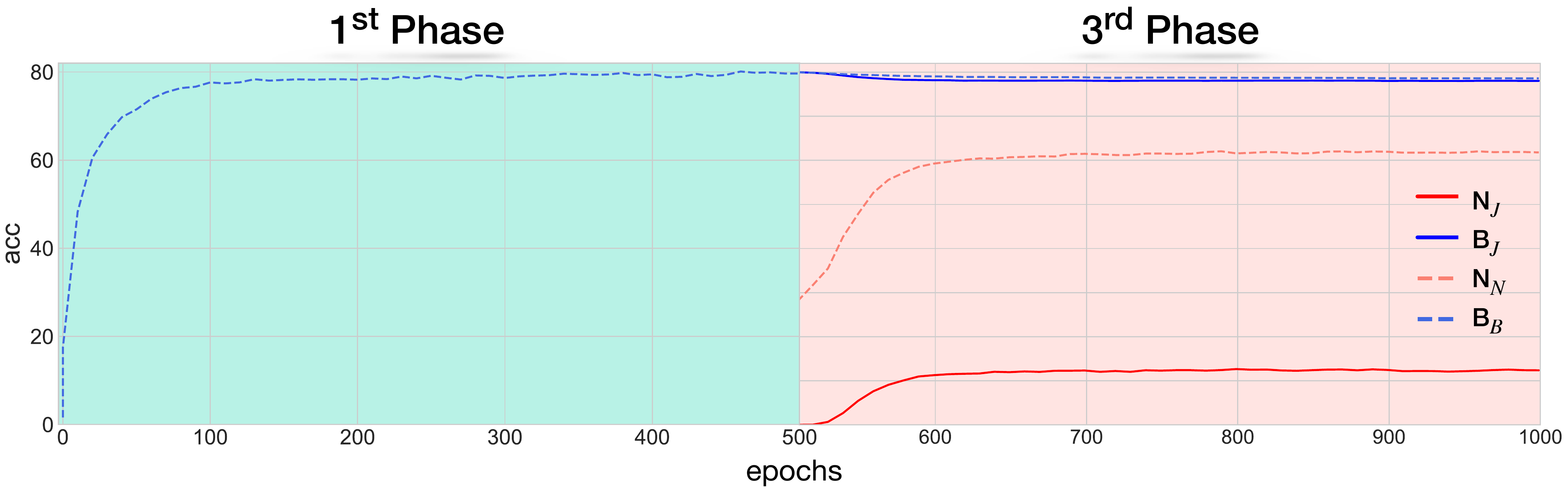}
\caption{We skip the second phase in our three-phase framework and sequentially apply the first and the third phase instead. The red curve ($N_{J}$) shows that the performance of the model on novel samples in the joint space stays low and is not able to achieve high accuracy in the joint space without the second phase.}
\label{fig:2nd_skip}
\end{figure}

\myparagraph{Skip the second phase}
In the proposed work, we address the problem of generalized few-shot learning with a three-phase framework. During the second phase we target to improve \textit{novel class learning} and to mitigate \textit{catastrophic forgetting} of the base classes. In Fig.~\ref{fig:2nd_skip} we show the development of the performance when we skip the second phase and directly proceed with the third phase.
During the third, joint calibration phase, the training set consists of base (one sample per class) and all novel training samples.
The performance of the base classes in the joint space $B_{J}$ and the separate space $B_{B}$ stays at high level even with few training samples.
While the separate novel $N_{N}$ performance can reach high values during the third phase, novel class learning in the joint space suffers from strong bias towards base classes (red curve on the figure stays low). It shows that our second phase for explicit novel learning in the joint space gives a significant boost to the overall performance in the joint space. \\
\\

\myparagraph{Skip the second phase and keep batch ratio during the third phase} We further evaluate the performance of the model without the second phase but with the third phase adaptation. Specifically, we ensure consistent batch-wise ratio between novel and base classes during the third phase. In Table~\ref{table:batch_ratios} the results show not only better performance than trivially skipping the second phase but also outperform the previous state-of-the-art~\cite{ren2019incremental}. Our proposed three-phase framework performs better on the novel classes.
\input{batch_ratios}
\myparagraph{Interleave the second and the third phases}
In order to shed light onto the question if one should separate the second and the third phases as proposed, in Table~\ref{table:alterante_phases}, we instead interleave the second and the third phases. Particularly, we alternate training on novel classes only (for $X$ epochs) and balanced replay (for 1 epoch).  We use $X=10, 20, 30$. Phase alternation shows to be an effective
alternative compared to the consecutive execution that still performs best.
\input{alternate_phases}


\subsection{Number of epochs of the second phase}
\begin{figure}[ht]
\centering
\includegraphics[width=\linewidth]{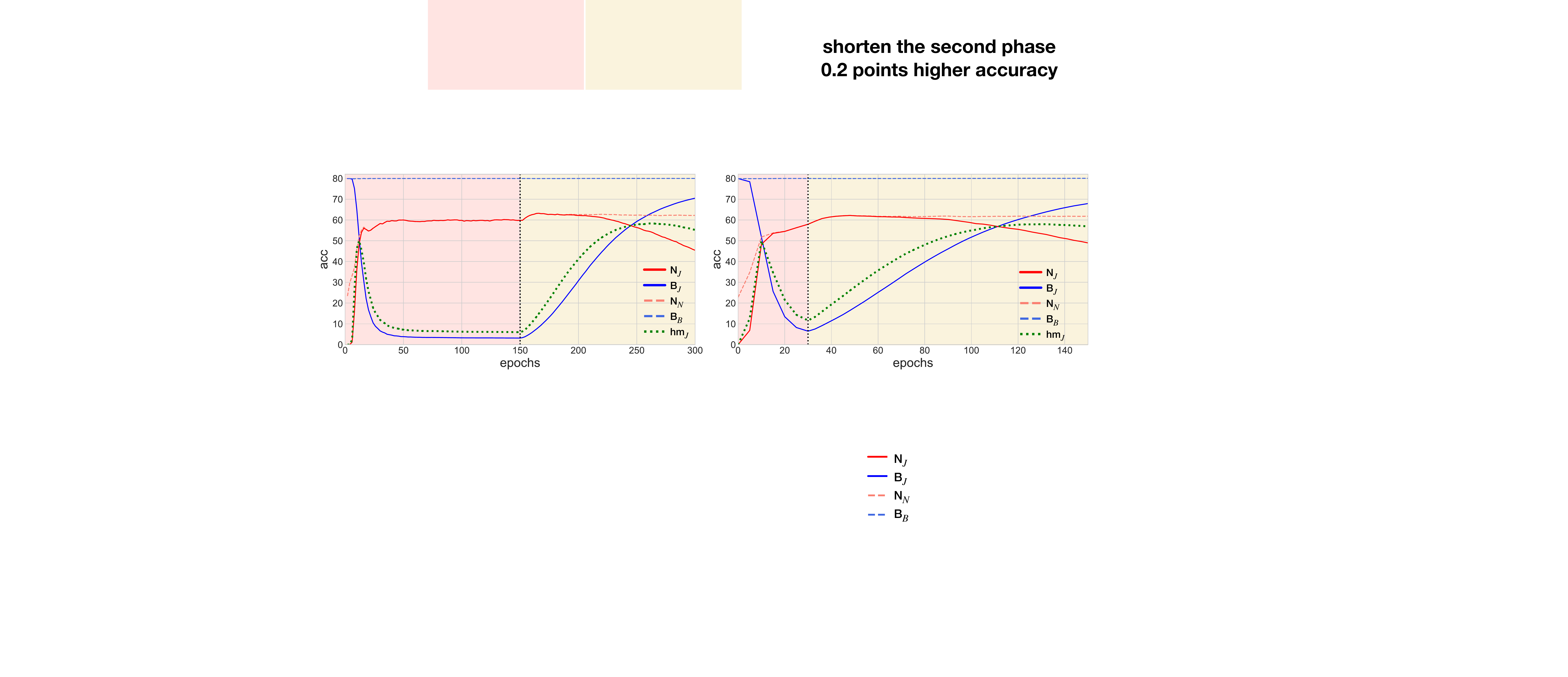}
\caption{Different duration of the second phase. The behaviour and the quantitative performance is the same. \textit{Left}: we use 150 epochs for the second phase; \textit{right}: we use 30 epochs for the second phase. }
\label{fig:2nd_30+}
\end{figure}

For the evaluation, we train the model for a fixed number of epochs during the second phase for all the datasets and setups. Fig.~\ref{fig:2nd_30+} shows similar behaviour when we apply smaller (30) number of epochs during the second phase and compare it to a longer second phase (150 epochs). Due to the negligible differences we use the initially chosen value that equals 150 epochs throughout the main paper.

\subsection{High $\lambda$ for the weight constraints}
In the main paper in Fig.4 we show the range of appropriate $\lambda$ to achieve good balanced performance for 5w1s and 5w5s setups.
We claim that $\lambda$ should not prevent novel class learning while preserving base performance in the base class space.
In Fig.4 we exclude too low $\lambda$ since empirically we found decrease in the performance on the base classes after the third phase.
Table~\ref{table:lambda_after_3_phase} presents the performance of the model with different $\lambda$ after the third phase.
Higher $\lambda$ helps to better preserve knowledge of the base classes while it hinders novel class learning in the joint space.
\input{lambda_after_3_phase}

\subsection{Impact of weight decay regularization}
\begin{figure}[ht]
\centering
\includegraphics[width=.8\linewidth]{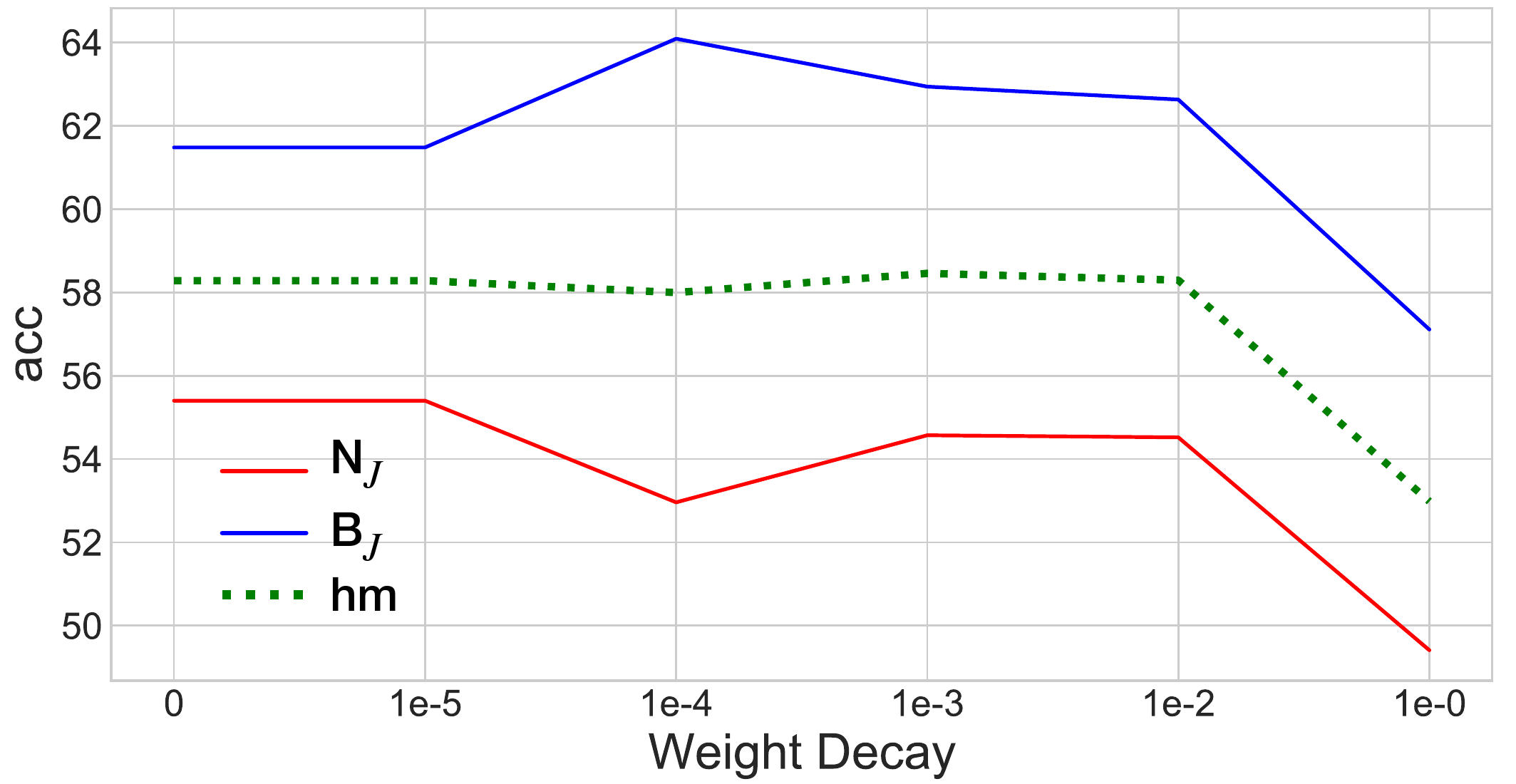}
\caption{Contribution of weight decay regularization and influence on the performance of the framework in term of $N_{/J}$, $B_{/J}$ and $hm_{/J}$. Results on mini-ImageNet, 5w1s, averaged over 100 episodes. }
\label{fig:weight_decay}
\end{figure}

While we apply constraints on the parameters of the model by applying $L_2^{WC}$, the question arises if and to which extend we would need standard weight decay as regularization on the model parameters. As shown in Fig.~\ref{fig:weight_decay}, while the contribution of the regularization term remains minor, it neither helps the performance nor harms.

\subsection{10w1s and 20w1s}
We evaluate our approach for additional setups to directly compare to knowledge preservation
methods, similarly as in Table 6 in the main paper. Table~\ref{table:distill_loss_rebuttal} confirms our finding that with little amount of data knowledge distillation (KD)~\cite{hinton2015distilling, rebuffi2017icarl}
performs worse than with $L_{2}^{WC}$ constraints.
\input{distill_loss_rebuttal}

\section{Extended implementation details}
\label{sec:impl_det_sup}
This section covers additional details of the implementation.
We first specify the exact augmentation for images and then discuss the evaluation for the generalized and incremental setups.
Our framework is built with PyTorch library and will be made publicly available. \\
\myparagraph{Augmentation}
For training on images we apply standard augmentation with random resizing followed by random cropping to the size 84x84 and random horizontal flip. We also use color jittering that allows to randomly change brightness, contrast, and saturation. For the test we first resize an image to the size of 92x92 and then apply a central 84x84 crop.\\
\myparagraph{Evaluation}
For each image and video dataset for testing we use 15 samples per class for both base and novel classes.
We train two parametric classifiers for base and novel classes respectively, to evaluate the performance in the joint space we concatenate the logit vectors (dimensionalilty $C_B$ and $C_N$ for base and novel classes accordingly) and predict the class by applying argmax operator over the concatenated output vector of dimensionality $C_B + C_N$.\\
\myparagraph{\textit{lim $\&$ unlim}}
For each dataset we conduct experiments with the two following setups: \textit{lim} denotes \textit{limited} access to the base training data during the third phase, whereas for \textit{unlim} we allow the model to have \textit{unlimited} access to the base training samples during the third phase. During the third phase we target to have balanced training set, thus the replay set consists of all $|C_N|\cdot K$ novel samples and $|C_B| \cdot K$ base samples. $K$ refers to the number of samples for each novel class, the notation corresponds the standard notation for few-shot learning~\cite{chen2021incremental, gidaris2018dynamic, kumar2019protogan}, e.g. 5w1s denotes 5 novel classes with 1 training sample per class, thus $K=1$. \textit{lim}: for each episode we draw at random the replay set only \textbf{once} before the third phase and reuse that replay set for each epoch for that episode. \textit{unlim}: for each episode we draw random base training samples for the replay set before each epoch anew.

\subsection{Generalized few-shot learning setup}
Episodic training is a common way to evaluate few-shot learning methods, further we detail the difference from the standard training protocol~\cite{gidaris2018dynamic, ren2019incremental, ye2020set2set}.
As before $C_B$ stands for base classes. Generally, the few-shot setup is formulated in a $N$-way $K$-shot notation ($C_N$ from the main paper equals to $N$ in this formulation), specifically each few-shot episode consists of $C_N$ novel classes with $K$ training samples per class. In the generalized setup each episode includes $C_B$ base classes for the classification along with $C_N$ novel classes.
Each dataset includes $T$ classes in the novel test set, usually $|T| \gg |C_N|$, e.g. in mini-ImageNet there are $T=24$ classes in the test set whereas we evaluate on 5w1s and 5w5s (for both cases $C_N=5$). To this end, following standard practice~\cite{gidaris2018dynamic, ren2019incremental, ye2020set2set} we evaluate the performance as average over 600 episodes such that for each episode we repeat:
\begin{enumerate}[label=\arabic*)]
    \item randomly draw $C_N$ novel classes from $T$
    \item randomly draw $K$ training samples per each class
    \item apply training framework to the current training data
    \item randomly draw 15 samples per class to test the framework
    \item reset the framework to initial state and clean train data
\end{enumerate}

\subsection{Incremental few-shot learning setup}
We refer to recent work~\cite{delange2021continual, prabhu2020gdumb} for an extensive overview and taxonomy on incremental (continual) learning.  In our work we aim at class-incremental learning where all seen classes should be classified in the joint space. Whilst another popular choice of continual learning is task-incremental learning with an objective to achieve high accuracy in the disjoint spaces, in our notation we could refer to this setup as having high $B_{/B}$ base performance and high $N_{/N}$ novel performance separately.
Such formulation of the task is easier than to achieve a joint balanced, high performance.
The usual way to evaluate class incremental learning~\cite{liu2020mnemonics, tao2020few, chen2021incremental} is to continuously measure performance of the model in the growing joint space.
The first task (sometimes called session) is to evaluate performance on the base classes $C_B$. Following few-shot $N$-way ($C_N$) $K$-shot notation, the second task increases the joint space by $|C_N|$ classes, the third task increases again by $|C_N|$ classes resulting in $|C_B| + 2 |C_N|$ classes and so on up to 9 tasks in mini-ImageNet, namely $|C_B| + 8 |C_N|$ classes. We follow~\cite{tao2020few, chen2021incremental} and use the same division into the tasks, training and test samples.

\subsection{UCF101 splits}
As mentioned in the main paper in Section 4, for UCF101 we have introduced a novel split. We observe that the $B_{/B}$ performance achieves almost $99$ points with the previously introduced split by Dwivedi~\cite{kumar2019protogan} that we report in Table~\ref{table:ucf_sota_50w1s_sup}. We do not change the division into base and novel classes but instead we filter out some videos that share the same group~\cite{Soomro2012UCF101} from train and test splits. When comparing the results of $B_{/B}$ from the previous split and our novel split we indeed see a drop in the performance indicating that the proposed split corresponds to a harder task. Subsequently, the performance $N_{/N}, B_{/J}$ and $N_{/J}$ also drops. We will make the novel split publicly available.

\section{Extended tables}
\label{sec:tabels_sup}
In Table~\ref{table:incremental_sota_9_task} we summarize Tables~\ref{table:incremental_sota_hm_sup},~\ref{table:incremental_sota_b_and_n_sup} and ~\ref{table:incremental_sota_both_sup} by reporting performance with different metrics after all 9 tasks for incremental few-shot learning. In the supplement we account \textit{base biased} and \textit{balanced hm} performance for our framework. \textit{base biased} stands for the performance of the model that shows higher accuracy on base classes, whereas \textit{balanced hm} indicates more balanced performance between the disjoint sets that we control by number of epochs for the third phase. The discrepancy is caused by the difference in the number of base and novel classes  ($|C_B| \gg |C_N|$) and the initial bias of the network towards base classes due to larger number of training samples and further knowledge preservation.
Therefore, in Table~\ref{table:incremental_sota_9_task} $J_{/J}$ performance mainly depends on the performance of the base classes $B_{/J}$, e.g. for Joint training method $B_{/J}$ and $J_{/J}$ show the highest 61.89 points and 43.38 points respectively among all other methods. All other methods that achieve high $B_{/J}$ performance (60.44 for IDLVQ, 59.64 for IW) accordingly reach high performance on the joint set of base and novel classes $J_{/J}$ (41.84 for IDLVQ, 41.26 for IW). At the same time these methods perform poorly on the novel samples that corresponds to $N_{/J}$ column in Table~\ref{table:incremental_sota_9_task} (15.62 for Joint, 13.94 for IDLVQ, 13.69 for IW). On the contrary, in our framework we explicitly address novel class learning in the joint $N_{/J}$ space via base-normalized cross entropy and, thus, we are able to surpass all the previous methods on novel classes by more than 10 points, we reach 27.65 points. Our \textit{base biased} model outperforms previous state-of-the-art models by large margin on novel classes $N_{/J}$, harmonic mean $hm_{/J}$, and sets a new benchmark for the joint classes $J_{/J}$. By \textit{balanced hm} we show that better balance can be achieved in terms of $N_{/J}$ and $hm_{/J}$, while $B_{/J}$ and accordingly $J_{/J}$ decreases.

Tables~\ref{table:tiered_sota_sup},~\ref{table:kinetics_sota_sup}, and~\ref{table:ucf_sota_50w1s_sup} are extension of Tables 2, 3, and 4 from the main paper respectively. For all the datasets we report additionally $N_{/N}, B_{/B}$, and $am_{/J}$.
\input{sota_incremental_sum_sup}
\input{sota_tiered_sup}
\input{sota_kinetics_sup}
\input{sota_ucf_sup}

\input{sota_incremental_sup}

%% file: batch_ratios.tex

\setlength\belowrulesep{1pt}
\setlength\aboverulesep{2pt}

\begin{table}[h]
\centering
\small
\tabcolsep=0.15cm
\begin{tabular}{ccccc}  
\toprule
batch size & per batch ratio & $N_{/J}$  & $B_{/J}$  & \multirow{2}{*}{$hm_{/J}$}  \\
$\# N + \# B$ & $N / B$& ($5/_{69}$) & ($64/_{69}$)  & \\
\midrule
$5+1$ & $83/17$ & 52.39 & 57.28 & 54.72  \\
$5+2$ & $71/29$ & 51.51 & 59.59 & 55.26 \\
$5+3$ & $63/37$ & 48.11 & 63.52 & 54.75 \\
\midrule
ANN~\cite{ren2019incremental} & - & 45.61 & 63.92 & 53.24 \\
LCwoF & 3 phases & 53.28 & 63.24 & 57.83  \\
\bottomrule
\end{tabular}

\caption{5w1s mini-ImageNet. No $3^{rd}$ phase, controlled batch ratio.
}

\label{table:batch_ratios}
\end{table}

%% file: alternate_phases.tex

\setlength\belowrulesep{1pt}
\setlength\aboverulesep{2pt}

\begin{table}[h]
\centering
\small
\tabcolsep=0.15cm
\begin{tabular}{ccccc}  
\toprule
epochs per period & period ratio & $N_{/J}$  & $B_{/J}$  & \multirow{2}{*}{$hm_{/J}$}  \\
$X(novel) + 1 (replay)$ & $N / B$& ($5/_{69}$) & ($64/_{69}$)  & \\
\midrule
$10+1$ & $46/54$ & 43.08 & 70.06 & 53.35  \\
$20+1$ & $62/38$ & 49.20 & 65.40 & 56.15 \\
$30+1$ & $71/29$ & 51.37 & 64.20 & 57.07 \\
\midrule
LCwoF & 3 phases & 53.28 & 63.24 & 57.83  \\
\bottomrule
\end{tabular}

\caption{5w1s mini-ImageNet. Interleave of the $2^{nd}$ and the $3^{rd}$ phases.
}

\label{table:alterante_phases}
\end{table}

%% file: lambda_after_3_phase.tex

\begin{table}[h]
\centering
\small
\tabcolsep=0.15cm
\begin{tabular}{c|ccc}  
\toprule
& \multicolumn{3}{c}{5w1s}  \\
$\lambda$  & $N_{/J}$  & $B_{/J}$  & $hm_{/J}$ \\
\midrule
1e+1 & 53.27 & 60.03 & 56.45 \\
1e+2 & 53.37 & 61.45 & 57.13  \\
5e+2 & 53.28 & 63.24 & 57.83 \\
5e+3 & 52.65 & 63.48 & 57.56 \\

\bottomrule
\end{tabular}

\caption{Performance of the model after the third phase . $\lambda$ stands for the importance weight of the $L_2^{WC}$ term. Higher $\lambda$, higher knowledge preservation, higher accuracy $B_{/J}$. Performance of the model after the second phase for the corresponding $\lambda$ can be found in Fig.4 in the main paper. Results on mini-ImageNet.
}
\label{table:lambda_after_3_phase}
\end{table}

%% file: distill_loss_rebuttal.tex

\setlength\belowrulesep{1pt}
\setlength\aboverulesep{2pt}

\begin{table}[ht]
\vspace{-3mm}
\centering
\small
\tabcolsep=0.15cm
\begin{tabular}{c|ccc|ccc}  
\toprule
& \multicolumn{3}{c|}{10w1s} & \multicolumn{3}{c}{20w1s} \\
\multirow{2}{*}{method}  & $N_{/J}$  & $B_{/J}$  & \multirow{2}{*}{$hm_{/J}$} & $N_{/J}$ & $B_{/J}$  & \multirow{2}{*}{$hm_{/J}$} \\
&  ($10/_{74}$) & ($64/_{74}$) &  & ($20/_{84}$) & ($64/_{84}$) &  \\

\midrule
$L_2^{WC}$ & 40.84 & 58.31 & 47.75 & 29.76 & 56.68 & 38.87 \\
$KD$ & 37.81 & 58.22 & 45.53 & 27.06 & 57.75 & 36.85 \\
\bottomrule
\end{tabular}

\caption{Comparison of knowledge preservation techniques on mini-ImageNet for setups 10w1s and 20w1s, with 10 and 20 novel classes correspondingly.
}

\label{table:distill_loss_rebuttal}
\end{table}

%% file: sota_incremental_sum_sup.tex

\begin{table*}[!ht]
\centering
\small
\tabcolsep=0.15cm
\begin{tabular}{c|cccc}  
\toprule
\multicolumn{5}{c}{mini-ImageNet} \\
method & $B_{/J}$ & $N_{/J}$ & $J_{/J}$ & $hm_{/J}$ \\
\midrule
FT$^\diamond$ & 1.46 & 1.36 & 1.42 & 1.40  \\
Joint$^\diamond$ & \textbf{61.89} & 15.62 & \textbf{43.38} & 24.95   \\
\midrule
iCaRL~\cite{rebuffi2017icarl}$^\diamond$ & 24.47 & 7.70 & 17.76 & 11.71  \\
UCIR~\cite{hou2019learning}$^\diamond$ & 21.57 & 8.27 & 16.25 & 11.96 \\
PN~\cite{snell2017prototypical}$^\diamond$ & 56.47 & 11.35 & 38.42 & 18.90  \\
ILVQ~\cite{xu2012ilvq}$^\diamond$ & 56.49 & 11.34 & 38.43 & 18.89  \\
SDC~\cite{yu2020sdc}$^\diamond$ & 59.87 & 13.30 & 41.24 & 21.77   \\
IW~\cite{qi2018imprinted}$^\diamond$ & 59.64 & 13.69 & 41.26 & 22.27 \\
IDLVQ~\cite{chen2021incremental} & \textbf{60.44} & 13.94 & 41.84 & 22.65  \\
TOPIC~\cite{tao2020few} & - & - & 24.42 & -   \\
\midrule
LCwoF (base biased) & 55.98 & 23.12 & \textbf{42.84} & 32.73  \\
LCwoF (balanced hm) & 47.73 & \textbf{27.65} & 39.70 & \textbf{35.02}  \\

\bottomrule
\end{tabular}

\caption{IFSL. Comparison to state-of-the-art on mini-ImageNet using metrics $B_{/J}$, $N_{/J}$, $J_{/J}$, $hm_{/J}$ and $am_{/J}$ after the last (9) task. $^\diamond$ indicates results copied from IDLVQ~\cite{chen2021incremental}.}
\label{table:incremental_sota_9_task}
\end{table*}

%% file: sota_tiered_sup.tex

\setlength\belowrulesep{0pt}
\setlength\aboverulesep{0pt}

\begin{table*}[!th]
\centering
\small
\tabcolsep=0.15cm
\begin{tabular}{c|cc|cc|cc||cc|cc|cc}  
\toprule
& \multicolumn{6}{c}{tiered-ImageNet 5w1s} & \multicolumn{6}{c}{tiered-ImageNet 5w5s} \\
\multirow{2}{*}{method} &  $N_{/N}$ & $B_{/B}$ & $N_{/J}$  & $B_{/J}$  & \multirow{2}{*}{$hm_{/J}$} & \multirow{2}{*}{$am_{/J}$} &  $N_{/N}$ & $B_{/B}$ & $N_{/J}$  & $B_{/J}$  & \multirow{2}{*}{$hm_{/J}$} & \multirow{2}{*}{$am_{/J}$} \\
 & ($5_{/5}$) & ($200_{/200}$) & ($5_{/205}$) & ($20_{/205}$) &  &  & ($5_{/5}$) & ($200_{/200}$) & ($5_{/205}$) & ($200_{/205}$) &  &  \\
 \midrule
  PN~\cite{snell2017prototypical}$^{\star}$ & - & - & - & - & - & 30.04 & - & - & - & - & - & 41.38  \\
   IW~\cite{qi2018imprinted}(i) & 60.88 & 70.19 & 44.95 & 62.53 & 52.30 & 53.74 & 79.26 & 70.25 & 71.85 & 56.11 & 63.01 & 63.98  \\
 DFSL~\cite{gidaris2018dynamic} (c)  & 59.52 & 47.53 & 47.32 & 36.10 & 40.96 & 41.71 & 75.89 & 47.98 & 67.94 & 39.08 & 49.61 & 53.51  \\
 AAN~\cite{ren2019incremental}(c)  & 61.37 & 62.44 & 54.39 & 55.85 & 55.11 & 55.12 & 77.91 & 62.36 & 72.09 & 57.76 & 64.13 & 64.93  \\
  AAN~\cite{ren2019incremental}(orig)  & - & - & - & - & - & 56.11 & - & - & - & - & - & 65.52  \\
 LCwoF (ours) \textit{lim}  & 64.71 & 70.55 & 57.13 & 60.39 & \textbf{58.71} & \textbf{58.76} & 79.72 & 70.58 & 69.05 & 63.44 & \textbf{66.12} & 66.25  \\
 LCwoF (ours) \textit{unlim}  & 64.67 & 70.59 & 57.54 & 60.09 & \textbf{58.78} & \textbf{58.82} & 80.02 & 70.59 & 70.20 & 63.01 & \textbf{66.41} & 66.61  \\
\bottomrule
\end{tabular}

\caption{Comparison to state-of-the-art on tiered-ImageNet 5w1s (left) and 5w5s (right) with ResNet backbone. \textit{lim} denotes limited access to base train samples during the third phase, for \textit{unlim} we do not apply such restrictions.  $^{\circ}$ indicates results copied from RGFSL~\cite{shi2019relational}, $^{\star}$ indicates results from AAN~\cite{ren2019incremental}, (c) denotes that we run available code on the corresponding data, (i) states for our re-implementation of the respective method, (orig) indicates original numbers from the respective paper.}
\label{table:tiered_sota_sup}
\vspace{-2mm}
\end{table*}


%% file: sota_kinetics_sup.tex

\setlength\belowrulesep{0pt}
\setlength\aboverulesep{0pt}

\begin{table*}[!th]
\centering
\small
\tabcolsep=0.15cm
\begin{tabular}{c|cc|cc|cc||cc|cc|cc}  
\toprule
& \multicolumn{6}{c}{mini-Kinetics 5w1s} & \multicolumn{6}{c}{mini-Kinetics 5w5s} \\
\multirow{2}{*}{method} &  $N_{/N}$ & $B_{/B}$ & $N_{/J}$  & $B_{/J}$  & \multirow{2}{*}{$hm_{/J}$} & \multirow{2}{*}{$am_{/J}$} &  $N_{/N}$ & $B_{/B}$ & $N_{/J}$  & $B_{/J}$  & \multirow{2}{*}{$hm_{/J}$} & \multirow{2}{*}{$am_{/J}$} \\
 & ($5_{/5}$) & ($64_{/64}$) & ($5_{/69}$) & ($64_{/69}$) &  &  & ($5_{/5}$) & ($64_{/64}$) & ($5_{/69}$) & ($64_{/69}$) &  &  \\
 \midrule
   IW~\cite{qi2018imprinted}(i) & 62.57 & 58.42 & 45.56 & 48.56 & 47.01 & 47.06 & 74.61 & 56.67 & 56.92 & 49.17 & 52.76 & 53.05  \\
 DFSL~\cite{gidaris2018dynamic} (c)  & 65.35 & 56.04 & 50.81 & 44.51 & 47.45 & 47.66 & 81.11 & 56.46 & 70.29 & 46.31 & 55.83 & 58.30  \\
 GFSV~\cite{xian2020generalized}  & - & - & 13.70 & 88.70 & 23.73 & 51.20 & - & - & 22.30 & 88.70 & 35.64 & 55.50  \\
 AAN~\cite{ren2019incremental}(c) & 59.36 & 57.99 & 46.13 & 35.96 & 40.41 & 41.05 & 76.83 & 59.49 & 56.99 & 43.21 & 49.15 & 46.18  \\
 LCwoF (ours) \textit{lim} & 55.97 & 64.84 & 47.51 & 50.84 & \textbf{49.12} & 49.18 & 74.76 & 65.06 & 63.65 & 54.55 & \textbf{58.75} & 59.10  \\
 LCwoF (ours) \textit{unlim}  & 55.39 & 65.01 & 46.26 & 51.94 & \textbf{48.93} & 49.10 & 73.77 & 65.18 & 65.40 & 52.70 & \textbf{58.37}  & 59.05  \\
\bottomrule
\end{tabular}

\caption{Comparison to state-of-the-art on mini-Kinetics 5w1s (left) and 5w5s (right) with 2-layers MLP backbone. \textit{lim} denotes limited access to base train samples during the third phase, for \textit{unlim} we do not apply such restrictions.  (c) denotes that we run available code on the corresponding data, (i) states for our re-implementation of the respective method.}
\label{table:kinetics_sota_sup}
\vspace{-2mm}
\end{table*}


%% file: sota_ucf_sup.tex

\begin{table*}[!t]
\centering
\small
\tabcolsep=0.15cm
\begin{tabular}{c|cc|cc|cc}  
\toprule
\multicolumn{7}{c}{UCF101 50w1s} \\
\multirow{2}{*}{method} & $N_{/N}$ & $B_{/B}$  & $N_{/J}$  & $B_{/J}$  & \multirow{2}{*}{$hm_{/J}$} & \multirow{2}{*}{$am_{/J}$} \\
 & ($50_{/50}$) & ($51_{/51}$) & ($50_{/101}$) & ($51_{/101}$) &  &  \\
 \midrule
 ProtoG~\cite{kumar2019protogan} & - & - & 52.30 & 75.30 & 61.73 & 63.80 \\
  LCwoF (ours) & 57.13 & 98.97 & 54.41 & 91.41 & 68.22 & 72.91 \\
 
 \midrule
 
 IW~\cite{qi2018imprinted} & 54.08 & 85.59 & 45.22 & 76.15 & 56.73 & 60.69 \\
 LCwoF (ours) \textit{lim} & 55.98 & 84.16 & 50.78 & 70.72 & 59.11 & 60.75 \\
 LCwoF (ours) \textit{unlim} & 54.35 & 82.33 & 49.12 & 69.98 & 57.72 & 59.55 \\

\bottomrule
\end{tabular}

\caption{Comparison to state-of-the-art on UCF101 50w1s with 2 layers MLP backbone on pre-extracted features. (i) states for our re-implementation of the respective method. \textit{lim} denotes limited access to base train samples during the third phase, for \textit{unlim} we do not apply such restrictions. Top: splits from ProtoG~\cite{kumar2019protogan}; bottom: original UCF101 train/test splits as in~\cite{Soomro2012UCF101}. }
\label{table:ucf_sota_50w1s_sup}
\end{table*}

%% file: sota_incremental_sup.tex

\begin{table*}[!t]
\centering
\small
\tabcolsep=0.15cm
\begin{tabular}{c|ccccccccc}  
\toprule
\multicolumn{10}{c}{mini-ImageNet} \\
method & 1 & 2 & 3 & 4 & 5 & 6 & 7 & 8 & 9 \\
$hm_{/J}$ & $60$ & $+5$ & $+10$ & $+15$ & $+20$ & $+25$ & $+30$ & $+35$ & $+40$  \\
\midrule
FT$^\diamond$ & - & 7.23 & 7.39 & 4.87 & 2.40 & 2.06 & 1.84 & 1.57 & 1.40  \\
Joint$^\diamond$ & - & 8.92 & 17.02 & 21.86 & 20.54 & 22.92 & 22.85 & 24.41 & 24.95  \\
\midrule
iCaRL~\cite{rebuffi2017icarl}$^\diamond$ & - & 8.45 & 13.86 & 14.92 & 13.00 & 14.06 & 12.74 & 12.16 & 11.71  \\
UCIR~\cite{hou2019learning}$^\diamond$ & - & 9.62 & 14.14 & 15.58 & 13.19 & 13.63 & 13.11 & 12.76 & 11.96  \\
PN~\cite{snell2017prototypical}$^\diamond$ & - & 9.76 & 14.72 & 16.78 & 19.09 & 20.06 & 19.37 & 18.98 & 18.90  \\
ILVQ~\cite{xu2012ilvq}$^\diamond$ & - & 9.66 & 16.08 & 17.78 & 20.05 & 20.35 & 19.64 & 19.06 & 18.89  \\
SDC~\cite{yu2020sdc}$^\diamond$ & - & 20.51 & 18.79 & 17.36 & 20.47 & 19.21 & 18.27 & 20.79 & 21.77  \\
IW~\cite{qi2018imprinted}$^\diamond$ & - & 25.32 & 20.45 & 22.62 & 25.48 & 22.54 & 20.66 & 21.27 & 22.27  \\
IDLVQ~\cite{chen2021incremental} & - & 21.69 & 20.44 & 21.98 & 25.19 & 22.99 & 20.82 & 21.56 & 22.65  \\
\midrule
LCwoF (base biased) & - & 25.56 & 30.59 & 27.29 & 28.08 & 29.91 & 27.97 & 30.30 & 32.73  \\
LCwoF (balanced hm) & - & \textbf{41.24} & \textbf{38.96} & \textbf{39.08} & \textbf{38.67} & \textbf{36.75} & \textbf{35.47} & \textbf{34.71} & \textbf{35.02}  \\

\bottomrule
\end{tabular}

\caption{IFSL. Comparison to state-of-the-art on mini-ImageNet based on harmonic mean metric between base and novel classes. $^\diamond$ indicates results copied from IDLVQ~\cite{chen2021incremental}. }
\label{table:incremental_sota_hm_sup}
\end{table*}


\begin{table*}[!t]
\centering
\small
\tabcolsep=0.15cm
\begin{tabular}{c|ccccccccc}  
\toprule
\multicolumn{10}{c}{mini-ImageNet}  \\
$B_{/J}$ ($60$) & 1 & 2 & 3 & 4 & 5 & 6 & 7 & 8 & 9 \\
\midrule
FT$^\diamond$ & 64.25 & 32.28 & 20.87 & 6.95 & 3.17 & 3.16 & 1.92 & 1.53 & 1.46  \\
Joint$^\diamond$ & 64.25 & 63.30 & 62.83 & 62.16 & \textbf{62.18} & \textbf{62.68} & \textbf{61.86} & \textbf{61.87} & \textbf{61.89}  \\
\midrule
iCaRL~\cite{rebuffi2017icarl}$^\diamond$ & 64.25 & 51.66 & 48.97 & 45.62 & 37.39 & 30.86 & 28.68 & 26.83 & 24.47  \\
UCIR~\cite{hou2019learning}$^\diamond$ & 64.25 & 52.87 & 50.16 & 44.78 & 37.48 & 28.75 & 25.58 & 22.97 & 21.57  \\
PN~\cite{snell2017prototypical}$^\diamond$ & 64.25 & 59.27 & 58.88 & 58.69 & 58.22 & 57.63 & 57.03 & 56.80 & 56.47  \\
ILVQ~\cite{xu2012ilvq}$^\diamond$ & 64.25 & 60.24 & 59.62 & 59.02 & 58.61 & 57.71 & 57.16 & 56.83 & 56.49  \\
SDC~\cite{yu2020sdc}$^\diamond$ & 64.62 & 63.58 & 62.78 & 61.12 & 60.29 & 59.37 & 59.05 & 59.97 & 59.87  \\
IW~\cite{qi2018imprinted}$^\diamond$ & 64.71 & 63.52 & 62.96 & 62.13 & 61.17 & 61.27 & 60.63 & 59.86 & 59.64  \\
IDLVQ~\cite{chen2021incremental} & \textbf{64.77} & \textbf{63.77} & \textbf{63.22} & \textbf{62.44} & 61.22 & 61.47 & 60.97 & 60.66 & 60.44  \\
\midrule
LCwoF (base biased) & 64.45 & 63.53 & 62.07 & 61.55 & 60.85 & 59.26 & 58.25 & 57.23 & 55.98  \\
LCwoF (balanced hm) & 64.45 & 57.33 & 53.31 & 52.87 & 51.38 & 48.25 & 47.60 & 47.51 & 47.73  \\

\midrule
\midrule
$N_{/J}$ & 1 & 2 & 3 & 4 & 5 & 6 & 7 & 8 & 9 \\
$\#cl$& - & $5$ & $10$ & $15$ & $20$ & $25$ & $30$ & $35$ & $40$  \\
\midrule
FT$^\diamond$ & - & 4.07 & 4.49 & 3.75 & 1.93 & 1.53 & 1.77 & 1.61 & 1.36  \\
Joint$^\diamond$ & - & 4.80 & 9.84 & 13.26 & 12.30 & 14.03 & 14.01 & 15.21 & 15.62  \\
\midrule
iCaRL~\cite{rebuffi2017icarl}$^\diamond$ & - & 4.60 & 8.09 & 8.92 & 7.87 & 9.10 & 8.19 & 7.86 & 7.70  \\
UCIR~\cite{hou2019learning}$^\diamond$ & - & 5.29 & 8.23 & 9.43 & 8.00 & 8.93 & 8.81 & 8.83 & 8.27  \\
PN~\cite{snell2017prototypical}$^\diamond$ & - & 5.32 & 8.41 & 9.79 & 11.42 & 12.14 & 11.67 & 11.39 & 11.35  \\
ILVQ~\cite{xu2012ilvq}$^\diamond$ & - & 5.25 & 9.29 & 10.47 & 12.09 & 12.35 & 11.86 & 11.45 & 11.34  \\
SDC~\cite{yu2020sdc}$^\diamond$ & - & 12.23 & 11.05 & 10.12 & 12.33 & 11.46 & 10.81 & 12.58 & 13.30  \\
IW~\cite{qi2018imprinted}$^\diamond$ & - & 15.81 & 12.21 & 13.83 & 16.09 & 13.81 & 12.45 & 12.93 & 13.69  \\
IDLVQ~\cite{chen2021incremental} & - & 13.07 & 12.19 & 13.34 & 15.86 & 14.14 & 12.55 & 13.11 & 13.94  \\
\midrule
LCwoF (base biased) & - & 16.00 & 20.30 & 17.53 & 18.25 & 20.00 & 18.40 & 20.60 & 23.12  \\
LCwoF (balanced hm) & - & \textbf{32.20} & \textbf{30.70} & \textbf{31.00} & \textbf{31.12} & \textbf{29.68} & \textbf{28.27} & \textbf{27.34} & \textbf{27.65}  \\

\bottomrule
\end{tabular}

\caption{IFSL. Comparison to state-of-the-art on mini-ImageNet. Top: performance of the base samples in the joint space after each task. Bottom: performance of the novel samples in the joint space after each novel task. $^\diamond$ indicates results copied from IDLVQ~\cite{chen2021incremental}.}
\label{table:incremental_sota_b_and_n_sup}
\end{table*}


\begin{table*}[!t]
\centering
\small
\tabcolsep=0.15cm
\begin{tabular}{c|ccccccccc}  
\toprule
\multicolumn{10}{c}{mini-ImageNet $\text{both}^{\uplus}$} \\
 & 1 & 2 & 3 & 4 & 5 & 6 & 7 & 8 & 9 \\
$J_{/J}$ & $60$ & $+5$ & $+10$ & $+15$ & $+20$ & $+25$ & $+30$ & $+35$ & $+40$  \\
\midrule
FT$^\diamond$ & 64.25 & 30.11 & 18.53 & 6.31 & 2.86 & 2.86 & 1.87 & 1.56 & 1.42  \\
Joint$^\diamond$ & 64.25 & 58.80 & 55.26 & 52.38 & 49.71 & 48.37 & 45.91 & 44.68 & 43.38  \\
\midrule
iCaRL~\cite{rebuffi2017icarl}$^\diamond$ & 64.25 & 48.04 & 43.13 & 38.28 & 30.01 & 24.46 & 21.85 & 19.84 & 17.76  \\
UCIR~\cite{hou2019learning}$^\diamond$ & 64.25 & 49.21 & 44.17 & 37.71 & 30.11 & 22.92 & 19.99 & 17.96 & 16.25  \\
PN~\cite{snell2017prototypical}$^\diamond$ & 64.25 & 55.12 & 51.67 & 48.91 & 46.52 & 44.25 & 41.91 & 40.07 & 38.42  \\
ILVQ~\cite{xu2012ilvq}$^\diamond$ & 64.25 & 56.01 & 52.43 & 49.31 & 46.98 & 44.37 & 42.06 & 40.11 & 38.43  \\
SDC~\cite{yu2020sdc}$^\diamond$ & 64.62 & 59.63 & 55.39 & 50.92 & 48.30 & 45.28 & 42.97 & 42.51 & 41.24  \\
IW~\cite{qi2018imprinted}$^\diamond$ & 64.71 & 59.85 & 55.71 & 52.47 & 49.90 & 47.31 & 44.57 & 42.57 & 41.26  \\
IDLVQ~\cite{chen2021incremental} & 64.77 & 59.87 & 55.93 & 52.62 & 49.88 & 47.55 & 44.83 & 43.14 & 41.84  \\
TOPIC~\cite{tao2020few} & 61.31 & 50.09 & 45.17 & 41.16 & 37.48 & 35.52 & 32.19 & 29.46 & 24.42  \\
\midrule
LCwoF (base biased) & 64.45 & \textbf{59.88} & \textbf{56.10} & \textbf{52.75} & \textbf{50.20} & \textbf{47.71} & \textbf{44.97} & \textbf{43.74} & \textbf{42.84}  \\
LCwoF (balanced hm) & 64.45 & 55.40 & 50.08 & 48.49 & 46.28 & 42.78 & 41.16 & 40.08 & 39.70  \\

\bottomrule
\end{tabular}

\caption{IFSL. Comparison to state-of-the-art on mini-ImageNet based on joint performance of base and novel samples in the joint space. $^\diamond$ indicates results copied from IDLVQ~\cite{chen2021incremental}.}
\label{table:incremental_sota_both_sup}
\end{table*}